\pdfoutput=1

\documentclass[11pt]{article}

\usepackage[final]{acl}

\usepackage{times}
\usepackage{latexsym}

\usepackage[T1]{fontenc}

\usepackage[utf8]{inputenc}

\usepackage{microtype}

\usepackage{inconsolata}

\usepackage{graphicx}

\usepackage{makecell}
\usepackage{microtype}
\usepackage{hyperref}
\usepackage{url}
\usepackage{booktabs}
\usepackage{multirow}%
\usepackage{amsmath,amssymb,amsfonts}%
\usepackage{amsthm}%
\usepackage{mathrsfs}%
\usepackage[title]{appendix}%
\usepackage[normalem]{ulem}
\usepackage{xcolor}%
\usepackage{xspace}
\usepackage{textcomp}%
\usepackage{manyfoot}%
\usepackage{booktabs}%
\usepackage{algorithm}%
\usepackage{algorithmicx}%
\usepackage{algpseudocode}%
\usepackage{listings}%
\usepackage{subcaption}
\usepackage{commath}
\usepackage{tabularray}
\usepackage{bbding}
\usepackage{pifont}
\usepackage{utfsym}
\usepackage{times}
\usepackage{wasysym}
\usepackage{fontawesome}
\usepackage[normalem]{ulem}
\usepackage{colortbl}

\newcommand{\ming}{\textsc{MedCare}\xspace}
\newcommand{\ka}{\textsc{KA}\xspace}
\newcommand{\na}{\textsc{NA}\xspace}
\newcommand{\da}{\textsc{DA}\xspace}
\newcommand{\knowledgeaggregator}{\textsc{Knowledge Aggregator}\xspace}
\newcommand{\noiseaggregator}{\textsc{Noise Aggregator}\xspace}
\newcommand{\downstreamalignment}{\textsc{Downstream Alignment}\xspace}

\title{\ming: Advancing Medical LLMs through Decoupling Clinical Alignment and Knowledge Aggregation}

\newcommand{\huatuogptt}{HuatuoGPT-\uppercase\expandafter{\romannumeral2}}

\author{Yusheng Liao\thanks{Equal contribution.}$^{,\spadesuit,\diamondsuit}$, 
Shuyang Jiang$^{*, \clubsuit,\diamondsuit}$,
Zhe Chen$^{\spadesuit,\diamondsuit}$, 
Yanfeng Wang$^{\spadesuit,\diamondsuit}$,
Yu~Wang\thanks{Corresponding Author}$^{,\spadesuit,\diamondsuit}$
 \\
  $^{\spadesuit}$Shanghai Jiao Tong University \\
  $^{\clubsuit}$Fudan University \\
  $^{\diamondsuit}$Shanghai Artificial Intelligence Laboratory \\
  \texttt{\{liao20160907,chenzhe2018,wangyanfeng622,yuwangsjtu\}@sjtu.edu.cn} \\
  \texttt{shuyangjiang23@m.fudan.edu.cn}
}

\begin{document}
\maketitle

\begin{abstract}
Large language models (LLMs) have shown substantial progress in natural language understanding and generation, proving valuable especially in the medical field. Despite advancements, challenges persist due to the complexity and diversity inherent in medical tasks, which can be categorized as knowledge-intensive tasks and alignment-required tasks. Previous approaches either ignore the latter task or focus on a minority of tasks and hence lose generalization. To address these drawbacks, we propose a progressive fine-tuning pipeline. This pipeline employs a \knowledgeaggregator and a \noiseaggregator to encode diverse knowledge in the first stage and filter out detrimental information. In the second stage, we drop the \noiseaggregator to avoid the interference of suboptimal representation and leverage an additional alignment module optimized towards an orthogonal direction to the knowledge space to mitigate knowledge forgetting. Based on this two-stage paradigm, we proposed a \uline{\textbf{Med}}ical LLM through decoupling \uline{\textbf{C}}linical \uline{\textbf{A}}lignment and Knowledge Agg\uline{\textbf{re}}gation~(\ming), which is designed to achieve promising performance on over 20 medical tasks, as well as results on specific medical alignment tasks. Various model sizes of \ming (1.8B, 7B, 14B) all demonstrate significant improvements over existing models with similar model sizes. Our code and datasets are available at \url{https://github.com/BlueZeros/MedCare}.
\end{abstract}

\section{Introduction}


Large Language Models~(LLMs)~\citep{elmohamed,metaIntroducingMeta,singhal2023large} have made significant strides in the realms of natural language generation and thereby finding extensive applications across a broad spectrum of disciplines~\citep{LAWGPT-zh, deng2024k2}. Among these, the medical field has attracted considerable attention due to its importance and demand. Much effort has been dedicated to researching and developing specialized LLMs for this domain~\citep{li2023chatdoctor,wang2023huatuo,singhal2023large}. 

\begin{figure}[t]
    \centering
    \includegraphics[width=0.9\linewidth]{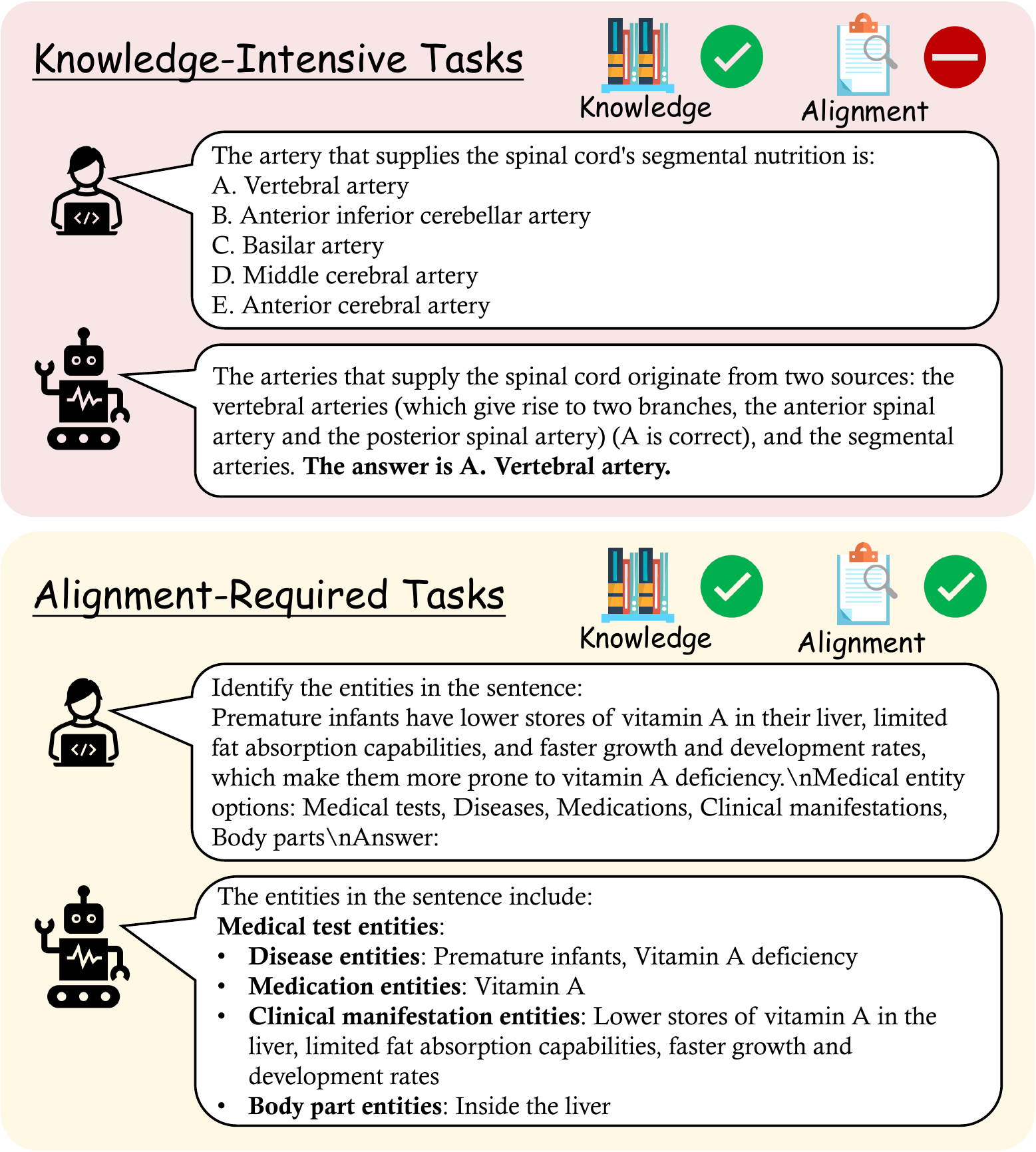}
    \caption{Examples of two types of medical tasks. Knowledge-intensive tasks require models to possess sufficient knowledge, whereas alignment-required tasks additionally necessitate the model to meet specific requirements criteria.}
    \label{fig:task_case}
\end{figure}

Although progress has been made, LLMs still face challenges due to the complexity and diversity of medical tasks. This is particularly evident when data is intentionally structured to represent a diverse set of tasks~\citep{chung2022scaling, wei2021finetuned, sanh2021multitask}.
We categorize medical tasks into two types: knowledge-intensive tasks and alignment-required tasks. Knowledge-intensive tasks, such as medical question answering~\citep{zhang2018medical} and medical dialogues~\citep{liu2022meddg,zhao2022medical}, primarily test the model's understanding of internal medical knowledge. Conversely, alignment-required tasks, like clinical terminology standardization~\citep{zhang2022cblue} and medical entity recognition~\citep{hongying2021building}, demand not only medical knowledge but also strict adherence to output formats. Figure~\ref{fig:task_case} illustrates these two categories. In general scenarios, like medical question-answering or dialogue, the LLMs only need to answer the user's questions correctly. However, tasks in clinical applications like report generation or situations usually require the model to produce a correct response and adhere to specific formats. 


Previous medical LLMs, such as HuatuoGPT~\citep{wang2023huatuo}, \huatuogptt~\citep{chen2023huatuogpt}, Med-PaLM~\citep{singhal2023large} and BioMistral~\citep{labrak2024biomistral}, have primarily focused on enhancing the encoding of medical knowledge in LLMs, neglecting the requirements of alignment-required tasks~\cite{van2024adapted}.
On the other hand, alignment-oriented methods~\cite {liu2023moelora,cai2023regemr,wang2023huatuo} that perform alignment through fine-tuning often suffer from an "alignment tax". This results in knowledge forgetting and performance drops in knowledge-intensive tasks~\citep{lin2023speciality,gekhman2024does}.
These issues limit the practical use of LLMs in healthcare.

\begin{figure*}[tbp]
    \centering
    \includegraphics[width=.9\linewidth]{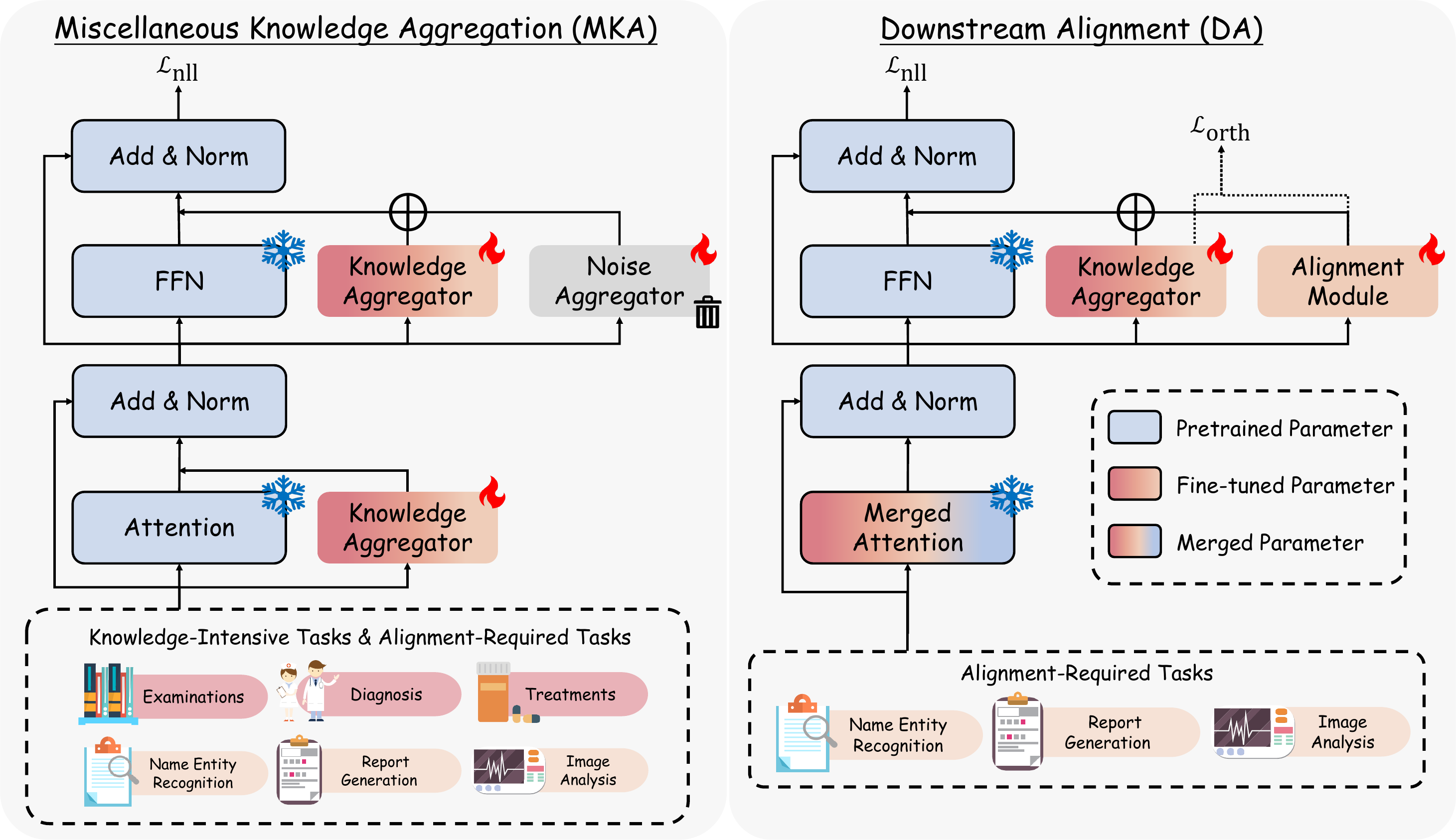}
    \caption{Overview of the proposed \ming.  In the MKA stage, \ming encodes advantageous knowledge and noisy contents with \knowledgeaggregator and \noiseaggregator from both types of tasks, respectively. The updated \noiseaggregator is removed to avoid the knowledge disruption. In the DA stage, an additional alignment module and orthogonal regularization are introduced to cater to the requirements of the alignment tasks.}
    \label{fig:ming}
\end{figure*}

To create a more practical medical LLM, in this paper we propose a novel training pipeline consisting of two progressive fine-tuning stages: miscellaneous knowledge aggregation~(MKA) and downstream alignment~(\da).
In the first stage, we introduce two modules, \knowledgeaggregator~(\ka) and \noiseaggregator~(\na), to encode advantageous knowledge and noisy contents, respectively.
The \noiseaggregator is asymmetric with respect to \knowledgeaggregator, which explicitly induces \knowledgeaggregator to perform multi-task knowledge extraction.
To avoid catastrophically parameterized knowledge disruption, we innovatively remove the updated \noiseaggregator but retain only the \knowledgeaggregator after training.
Following this stage, we introduce an alignment module to cater to the downstream alignment requirements from the specific task.
We add an orthogonal regularization term to ensure non-overlapping between the optimization space of alignment and knowledge space in the first stage.
Built upon this pipeline, we propose \ming with three sizes~(1.8B, 7B, 14B), a specialized LLM tailored for both knowledge-intensive tasks and alignment-required tasks, derived from the Qwen1.5 series.
Figure~\ref{fig:ming} provides a visual representation of \ming. 

In summary, our contributions are as follows:
\begin{itemize}
    \item We introduce a taxonomy for medical tasks, which divides all tasks into knowledge-intensive tasks and alignment-required tasks. This taxonomy reinforces the practical usability requirements for medical LLMs.
    
    \item We introduce a two-stage fine-tuning pipeline that balances knowledge maintenance and downstream alignment. This approach not only encodes knowledge without disruption but also adapts to specific tasks with minimal knowledge forgetting.
    
    \item Based on the progressive pipeline, we introduce \ming, a medical LLM to simultaneously encode massive knowledge and align with practical requirements under the medical multi-task taxonomy. To our knowledge, \ming is the first LLM to effectively handle such a wide spectrum of tasks in the medical domain with few alignment taxes.
    
    \item We conduct extensive experiments on over 20 knowledge-intensive and alignment-required tasks, benchmarking \ming against a diverse array of established language models. Our results demonstrate the superior performance of \ming, affirming its effectiveness for both genres of tasks.
\end{itemize}

\section{Preliminaries}
\paragraph{Low-rank Adaptation~(LoRA)}
Given an input sequence $\mathbf{X}\in\mathbb{R}^{m\times d}$, LoRA~\citep{hu2021lora} proves that the update of original linear layers $\Delta W$ in large language models is of low-rank, and can be decomposed into the multiplication of two compact matrices $AB$.
The pretrained weight $W$ is frozen in the training phase and does not undergo gradient updates, while $A$ and $B$ are trainable parameters and contribute together to the forward pass:
\begin{equation}
\label{lora}
    \mathbf{H}=\mathbf{X}W+ \mathbf{X}\Delta W = \mathbf{X}W + \frac{\alpha}{r}\mathbf{X}AB
\end{equation}
where $W\in\mathbb{R}^{d\times d'},A\in\mathbb{R}^{d\times r},B\in\mathbb{R}^{r\times d'}$ and $r\ll d,d'$. $\mathbf{H}$ is the processed output. $\alpha$ is the scaling factor. Without loss of generality, we omit the layer index for the following formula.
At the beginning of training, $B$ is initialized to an all-zero matrix and $A$ uses Gaussian initialization to make sure that the product $AB$ is zero at initialization.

\paragraph{Mixture of LoRA}
The mixture of the LoRA~(MoLoRA) module is based on the Feed-forward Network~(FFN) module in the LLMs since experts of FFN can store diverse task knowledge~\citep{geva2020transformer} from multi-task learning.
In the popular LLaMA-style FFN, the input $\mathbf{X}$ is computed using an SiLU~\citep{elfwing2018sigmoid} gate as follow:
\begin{equation}
    \mathbf{H}=(\mathbf{X}W_u\cdot\mathrm{silu}(\mathbf{X}W_g))W_d
\end{equation}
where $W_g\in\mathbb{R}^{d\times d'},W_u\in\mathbb{R}^{d\times d'},W_d\in\mathbb{R}^{d'\times d}$, and $d'=8/3d$.
In the LoRA forward passes, each linear layer $W\in\{W_d,W_u,W_g\}$ is updated using the LoRA module described in Eq.\ref{lora}.
Consider an MoLoRA module with $E$ experts, and each expert is denoted as $\{E_i\}_{i=1}^E$, the forward pass of the linear layer is formulated as:
\begin{align}
    \mathbf{H}&=\mathbf{X}W+ \mathbf{X}\Delta W\\
    \label{molora}
     \mathbf{X} \Delta W &= \sum_{i=1}^K G(\mathbf{X})_i E_{i}(\mathbf{X})
\end{align}
where $G(\cdot)=\mathrm{softmax}(\mathbf{X}W_r),W_r\in\mathbb{R}^{d\times E}$ is the router in MoLoRA with top-$K$ selection which holds the top-$K$ affinity with respect to current input $\mathbf{X}$.
Assuming each LoRA expert shares the same rank $r$ and $\alpha$ and we obtain the weight $w_i$ from the router $G(\cdot)$ as $w_i=G(\mathbf{X})_i$, the overall output of the MoLoRA linear layer can be formulated as:
\begin{equation}
    \mathbf{h}=W\mathbf{x}+\frac{\alpha}{r}\sum_{i=1}^K w_i \cdot  A_i B_i \mathbf{x}
\end{equation}
where $\mathbf{x},\mathbf{h}\in\mathbb{R}^{d}$ is any token representation of $\mathbf{X}$ and the frozen $W$ inherits from the pre-trained linear weight.

\section{\ming}
In this paper, we categorize the medical task into two primary genres: knowledge-intensive tasks and alignment-required tasks. Knowledge-intensive tasks require LLMs to seamlessly transfer both medical knowledge and their general domain reasoning capabilities into the medical domain. Models leverage the internal broad pre-trained knowledge without significantly disrupting the pretrained distribution, to fulfill domain-specific medical reasoning tasks. In contrast, alignment-required tasks pose a greater challenge. These tasks typically deviate more substantially from the model's original training paradigms, which can lead to hallucinations~\citep{gekhman2024does} and the aliasing of pre-trained knowledge~\citep{dou2023loramoe}.
To address this challenge, we propose \ming, a method designed to simultaneously acquire the necessary knowledge for the general adaptation of LLMs and to achieve specialized alignment for specific tasks. The fine-tuning process of \ming comprises two stages: miscellaneous knowledge aggregation (MKA; \S\ref{sec: knowledge_import}) and downstream alignment (DA; \S\ref{sec: downstream_alignment}). The training procedure for \ming is illustrated in Figure~\ref{fig:ming}.
\vspace{-0.6em}

\subsection{Problem Formulation}

In the MKA stage, the dataset $\mathcal{D}_{\text{MKA}}$ comprises $N$ samples $\{\mathbf{x}_i, \mathbf{y}_i\}_{i=1}^N$, derived from tasks that are both knowledge-intensive and require alignment. Here, $\mathbf{x}_i$ and $\mathbf{y}_i$ represent the input and target for the models, respectively. LLMs assimilate diverse knowledge from these tasks to augment their knowledge and reasoning capabilities. Subsequently, these LLMs are exposed to a specific subset of the alignment-required data $\mathcal{D}_{\rm DA}$, which is necessary for learning the intricacies of various alignment-required tasks, including named entity recognition with specificity and report analysis that demands explanatory detail.

\subsection{Miscellaneous Knowledge Aggregation}
\label{sec: knowledge_import}

In this stage, the models acquire common medical knowledge from both types of tasks. To absorb miscellaneous knowledge into single LLMs without interference, we adopt the MoLoRA structure~\citep{li2024mixlora,su2024mixture,feng-etal-2024-mixture-loras,luo2024moelora} as the \noiseaggregator and introduce shared experts as the \knowledgeaggregator to further circumvent low generalization due to the high specialization of each expert~\citep{gou2023mixture} and parameter redundancy~\citep{dai2024deepseekmoe} on the FFN module of the models. 
Through backpropagation, the \ka acquires common knowledge from multiple datasets, while \na learns the distinct alignment-required tasks. 
Consequently, the \ka absorbs common knowledge encoded in each task, while the \na learns the interfering alignment requirements.
\begin{align}
    \mathbf{H}&=\mathbf{X}W+\mathrm{KA}(\mathbf{X})+\mathrm{NA}(\mathbf{X}) \\
    &=\mathbf{X}W+\mathrm{LoRA}_s(\mathbf{X})+\mathrm{MoLoRA}(\mathbf{X})
\end{align}
where $W$ denotes any weight matrix in FFN modules. The rank of NA is $r$, and the rank of KA is $r'=sr$, where $s$ is the number of the share experts. For the attention module, we only add a vanilla LoRA module to each projection layer $W$ as Eq.\ref{lora} to guarantee sufficient optimization space.
As most task knowledge is encoded in the common knowledge aggregator, we discard the separate knowledge aggregator after this training stage. In other words, we introduce redundant parameters to avoid erroneous leverage of task-agnostic knowledge through decoupling the KA and NA modules, employing only the \ka module:
\begin{equation}
    \mathbf{H}=\mathbf{X}W+\mathrm{KA}(\mathbf{X})
\end{equation}
While discarding the separate KA prevents the model's knowledge from being disturbed, it also compromises the model's alignment capabilities, thus leading to suboptimal performance on alignment-required tasks. Therefore, a second stage of training is necessary to enable the model to learn the requirements of alignment tasks while preserving its knowledge reasoning capabilities.

\begin{table*}[t]
\centering
\resizebox{\textwidth}{!}{%
\begin{tabular}{lccccccccc}
\toprule
\textbf{Model} & \textbf{MedQA} & \textbf{MMB.} & \textbf{CMB} & \textbf{CMExam} & \textbf{CMMLU} & \textbf{CEval} & \textbf{PLE Pha} & \textbf{PLE TCM} & \textbf{Avg.} \\
\midrule
Llama3-8B & 59.40 & 63.78 & 41.63 & 44.99 & 51.40 & 53.66 & 38.33 & 33.54 & 48.34 \\
ChatGLM3-6B & 44.51 & 51.34 & 39.81 & 43.21 & 46.97 & 48.80 & 34.60 & 32.90 & 42.77 \\
Baichuan2-7B & 45.97 & 52.39 & 46.33 & 50.48 & 50.74 & 51.47 & 44.60 & 42.10 & 48.01 \\
Baichuan2-13B & 49.42 & 56.71 & 50.87 & 54.90 & 52.95 & 58.67 & 44.20 & 41.70 & 51.18 \\
Qwen1.5-7B & 74.46 & 78.58 & 61.33 & 58.24 & 66.77 & 68.29 & 32.29 & 29.17 & 58.64 \\
Qwen1.5-14B & \textbf{81.93} & 84.33 & 64.33 & 57.79 & 68.76 & 78.05 & 48.33 & 38.33 & 65.23 \\
ChatGPT & 37.51 & 40.08 & 43.26 & 46.51 & 50.37 & 48.80 & 41.20 & 31.20 & 42.81 \\
HuatuoGPT-II-7B & 59.22 & 62.03 & 60.39 & 65.81 & 59.08 & 62.40 & 47.70 & 47.50 & 58.02 \\
HuatuoGPT-II-13B & 75.77 & 78.69 & 63.34 & 68.98 & 61.45 & 64.00 & 52.90 & 51.60 & 64.59 \\
\midrule
\ming-1.8B & 56.80 & 65.62 & 43.04 & 47.83 & 48.82 & 51.22 & 38.13 & 35.21 & 48.33 \\
\hspace{1em} $\mapsto$ \emph{w/o} DA & 61.70 & 64.62 & 49.28 & 52.21 & 51.18 & 60.98 & 36.04 & 39.58 & 51.95 \\
\ming-7B & 76.77 & 80.79 & 60.13 & 65.33 & 64.33 & 65.85 & 47.29 & 52.08 & 64.07 \\
\hspace{1em} $\mapsto$ \emph{w/o} DA & 75.16 & 79.36 & 61.60 & 66.85 & 66.25 & 70.73 & 50.83 & 53.33 & 65.51 \\
\ming-14B & 81.44 & \textbf{84.76} & 64.17 & 68.74 & \textbf{71.57} & 78.05 & 54.17 & \textbf{54.58} & 69.69 \\
\hspace{1em} $\mapsto$ \emph{w/o} DA & 77.41 & 83.36 & \textbf{66.79} & \textbf{71.77} & 69.42 & \textbf{82.93} & \textbf{57.29} & 54.17 & \textbf{70.39} \\
\bottomrule
\end{tabular}%
}
\caption{The results on medical knowledge exams. The results of the MedQA are evaluated on the Chinese Mainland subset. `MMB.' indicates the Chinese subset of the MMedBench. `PLE Pha' and `PLE TCM' indicate the Pharmacy and Traditional Chinese Medicine tracks of the 2023 Chinese National Pharmacist Licensure Examination. Note that for the general benchmarks, CMMLU and CEval, we only chose questions related to the medical domain.}
\label{tab: knowledge experiments}
\end{table*}

\subsection{Downstream Alignment}
\label{sec: downstream_alignment}
In this stage, \ming merges back the LoRA weights of self-attention into the pretrained model and freezes the self-attention module to ensure the instruction-following ability of LLMs~\citep{wu2023language}. We introduce an additional alignment LoRA module $\mathrm{Align}(\cdot)$ in the FFN module to acquire specific alignment knowledge from the alignment dataset, the forward pass of linear layers of FFN can be formulated as:
\begin{align}
\label{alignment}
\mathbf{H}&=\mathbf{X}W+\mathrm{KA}(\mathbf{X})+\mathrm{Align}(\mathbf{X}) 
\end{align}
Following \citet{wang-etal-2023-orthogonal}, we introduce an orthogonal loss to ensure that the new alignment task is learned in a direction orthogonal to the original knowledge task.
For the LoRA weights of \ka $\{A^p_k, B^p_k\}_{p=1}^{P}$ and Align $\{A^p_d, B^p_d\}_{p=1}^{P}$, where $P$ is the total number of modules, we achieve the orthogonal subspaces of the alignment stage as:
\begin{align}
    \arg\min_{A^p_d} O^p_{k,d}={A^p_k}^\top A^p_d
\end{align}
Based on the subspace learning, the learning objective of this stage is expressed as
\begin{align}
    \mathcal{L}&=\mathcal{L}_{\mathrm{nll}}+\lambda\mathcal{L}_{\mathrm{orth}} \\
    &=-\sum_{\mathbf{x},\mathbf{y}\in\mathcal{D}_{\mathrm{DA}}}\log p_{\theta}(\mathbf{y}|\mathbf{x})+\lambda \sum_{p=1}^{P}\abs{O^p_{k,d}}
\end{align}
where $\abs{O_{k,d}}$ denotes the sum of the absolute value of each entry of $O_{k,d}$, and $\lambda$ is a hyperparameter to control the weights of orthogonal loss.
During the training process, we do not follow \citet{wang-etal-2023-orthogonal} to fix the knowledge aggregator module since they can further acquire the specific knowledge that is lacking in the previous learning stage.
After training, we can merge the updates of the knowledge aggregator and alignment module into the pretrained weights $W$ to avoid the increased inference latency and GPU overhead:
\begin{align}
    W' = W + \frac{\alpha}{r'}A_k B_k + \frac{\alpha}{r'}A_d B_d
\end{align}

\section{Experiments}
\ming is built upon the Qwen1.5-Chat~\footnote{\url{https://github.com/QwenLM/Qwen1.5}} series. For knowledge-intensive tasks, we adopt the Chinese Mainland test set of \textbf{MedQA}~\citep{jin2021disease}, the Chinese subset of \textbf{MMedBench}~\citep{qiu2024towards}, and two comprehensive Chinese medical exam datasets, \textbf{CMB}~\citep{wang2023cmb} and \textbf{CMExam}~\citep{liu2024benchmarking}. We also collect the medical parts of the general benchmarks, including \textbf{CMMLU}~\citep{li2023cmmlu} and \textbf{CEval}~\citep{huang2023ceval}. 
Besides, we test the performance of the models on the flash exam questions from the 2023 Chinese National Pharmacist Licensure Examination~(\textbf{PLE}) collected by~\citet{chen2023huatuogpt}. 
For the alignment-required tasks, we use \textbf{CBLUE}~\citep{zhang-etal-2022-cblue} and our newly developed Chinese Clinical Task Evaluation~(\textbf{CCTE}) dataset. More details of the experiment settings can be found in Appendix~\ref{appendix: experiments}.

\begin{figure*}[tbp]
    \centering
    \includegraphics[width=0.9\linewidth]{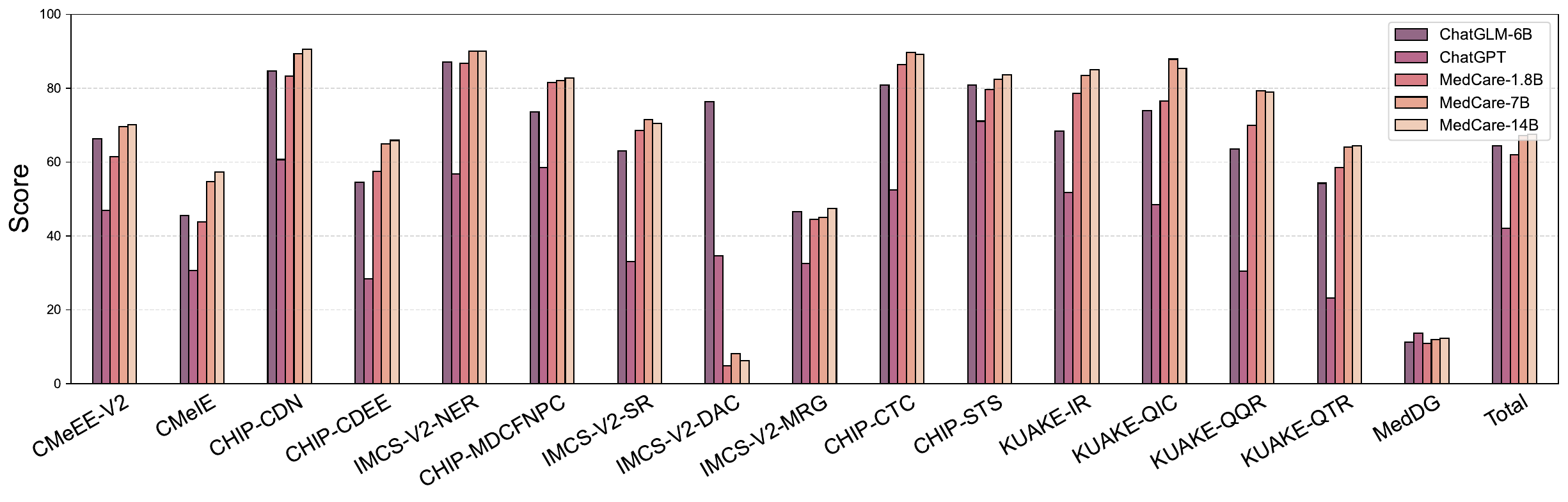}
    \caption{Results on 16 tasks in CBLUE. ChatGLM-6B is fine-tuned on the CBLUE dataset with LoRA and ChatGPT is augmented by in-context learning. The results are obtained from the official implementation of CBLUE.}
    \label{fig: cblue score}
\end{figure*}

\begin{table*}[tbp]
\centering
\resizebox{\textwidth}{!}{%
\begin{tabular}{l|ccccc|ccccc|ccccc|ccccc|c}
\toprule
 & \multicolumn{5}{c|}{\textbf{Report Generation}} & \multicolumn{5}{c|}{\textbf{Image Analysis}} & \multicolumn{5}{c|}{\textbf{Discharge Instruction}} & \multicolumn{5}{c|}{\textbf{Examination Education}} &  \\
\multirow{-2}{*}{\textbf{Model}} & \textbf{Flu.} & \textbf{Rel.} & \textbf{Com.} & \textbf{Pro.} & \textbf{Avg.} & \textbf{Flu.} & \textbf{Rel.} & \textbf{Com.} & \textbf{Pro.} & \textbf{Avg.} & \textbf{Flu.} & \textbf{Rel.} & \textbf{Com.} & \textbf{Pro.} & \textbf{Avg.} & \textbf{Flu.} & \textbf{Rel.} & \textbf{Com.} & \textbf{Pro.} & \textbf{Avg.} & \multirow{-2}{*}{\textbf{Avg.}} \\
\midrule
ChatGLM2-6B & 4.16 & 2.66 & 2.64 & 2.52 & 3.00 & 4.88 & 4.06 & 4.04 & 3.70 & 4.17 & 4.98 & 3.46 & 3.12 & 3.24 & 3.70 & 3.68 & 2.94 & 2.44 & 2.76 & 2.96 & 3.46 \\
ChatGLM3-6B & 4.52 & 2.78 & 2.66 & 2.52 & 3.12 & 4.98 & 4.16 & 4.20 & 3.94 & 4.32 & 5.00 & 3.34 & 3.24 & 3.04 & 3.66 & 4.52 & 2.88 & 2.14 & 2.56 & 3.03 & 3.53 \\
Baichuan2-7B & 4.88 & 3.36 & 3.30 & 3.40 & 3.74 & 5.00 & 4.40 & 4.40 & 4.02 & 4.46 & 5.00 & 3.96 & 3.84 & 3.82 & 4.16 & 3.98 & 2.74 & 1.68 & 2.20 & 2.65 & 3.75 \\
Baichuan2-13B & 4.94 & 3.42 & 3.40 & 3.46 & 3.81 & 5.00 & 4.42 & 4.42 & 4.22 & 4.52 & 5.00 & 3.92 & 3.92 & 3.88 & 4.18 & 4.56 & 2.88 & 1.40 & 2.18 & 2.76 & 3.81 \\
Qwen1.5-7B & 4.72 & 3.22 & 3.26 & 3.12 & 3.58 & 5.00 & 4.64 & 4.64 & 4.42 & 4.68 & 5.00 & 4.18 & 4.40 & 4.22 & 4.45 & 4.94 & 3.34 & 1.80 & 2.94 & 3.26 & 3.99 \\
Qwen1.5-14B & 4.86 & 3.54 & 3.50 & 3.50 & 3.85 & 5.00 & 4.66 & 4.66 & 4.44 & 4.69 & 5.00 & 4.10 & 4.16 & 4.10 & 4.34 & 4.80 & 3.28 & 1.68 & 2.56 & 3.08 & 3.99 \\
ChatGPT & 4.88 & 3.32 & 2.96 & 3.12 & 3.57 & 4.98 & 4.30 & 4.04 & 4.00 & 4.33 & 5.00 & 3.90 & 3.68 & 3.64 & 4.06 & 4.60 & 3.36 & 2.44 & 2.96 & 3.34 & 3.82 \\
HuatuoGPT-II-7B & 4.80 & 3.16 & 3.12 & 3.30 & 3.60 & 5.00 & 4.22 & 4.26 & 4.04 & 4.38 & 5.00 & 3.96 & 3.90 & 3.88 & 4.19 & 5.00 & 4.08 & 3.88 & 3.90 & 4.22 & 4.09 \\
HuatuoGPT-II-13B & 4.78 & 3.22 & 3.16 & 3.40 & 3.64 & 5.00 & 4.34 & 4.30 & 4.14 & 4.45 & 5.00 & 3.90 & 3.88 & 3.84 & 4.16 & 4.90 & 4.30 & 4.02 & 3.96 & 4.30 & 4.13 \\
\midrule
\ming-1.8B & 5.00 & 3.66 & 3.54 & 3.78 & 4.00 & 4.98 & 3.92 & 4.00 & 3.86 & 4.19 & 5.00 & 4.18 & 4.24 & 4.12 & 4.39 & 4.72 & 4.20 & 3.90 & 3.88 & 4.18 & 4.19 \\
\ming-7B & 5.00 & 3.78 & 3.68 & 3.80 & 4.07 & 5.00 & 4.12 & 4.24 & 4.12 & 4.37 & 5.00 & 4.36 & 4.42 & 4.38 & 4.54 & 4.96 & 4.42 & 4.40 & 4.34 & 4.53 & 4.38 \\
\ming-14B & 4.98 & 3.74 & 3.62 & 3.70 & 4.01 & 5.00 & 4.14 & 4.24 & 4.10 & 4.37 & 5.00 & 4.46 & 4.46 & 4.40 & 4.58 & 4.94 & 4.54 & 4.48 & 4.44 & 4.60 & \textbf{4.39} \\
\bottomrule
\end{tabular}%
}
\caption{Results on medical alignment task CCTE. `Flu.' indicates `Fluency', `Rel.' indicates `Relevance', `Com.' indicates `Completeness', and `Pro.' indicates `Proficiency'. The maximum value of all scores is 5.}
\label{tab: ccte results}
\end{table*}




\subsection{Results on Knowledge-Intensive Tasks}

We utilized eight distinct datasets to rigorously assess the medical knowledge capabilities of the proposed model \ming for comprehensive evaluation. As detailed in Table~\ref{tab: knowledge experiments}, \ming has demonstrated exceptional proficiency in medical knowledge, particularly notable in the \ming-14B, which has outperformed both ChatGPT and the leading open-source medical model, HuatuoGPT-II~(13B). Remarkably, the \ming-1.8B model matches the performance of ChatGLM3-6b, while \ming-7B achieves comparable results to the HuatuoGPT-II (13B). Besides, without the \downstreamalignment, \ming achieves better performance. 

To mitigate potential biases caused by data leakage, we further evaluated our models on the latest 2023 Chinese National Pharmacist Licensure Examinations (PLE)~\citep{chen2023huatuogpt}, with the detailed results shown in Table~\ref{tab: ple detail results}. This assessment confirms that \ming continues to outpace the competition, with \ming-14B \emph{w/o} DA surpassing the performance of GPT-4. These findings underscore \ming's superior utilization of medical knowledge, achieving higher expertise levels with fewer parameters. 

\subsection{Results on Alignment-Required Tasks}
To evaluate the medical alignment ability of the models, we tested CBLUE and CCTE with 20 distinct alignment-required tasks. CBLUE requires models to generate outputs in a prescribed format. A higher score indicates the model aligns more closely with the task requirements. As depicted in Figure~\ref{fig: cblue score}, the results indicate that even when augmented by In-Context Learning (ICL), ChatGPT still fails to fulfill such specific format requirements.
Besides, \ming-7B surpasses the established CBLUE baseline ChatGLM on nearly all the tasks with an average of more than 3 points. Since CBLUE strictly requires the output format, we mainly evaluate model alignment capabilities through ablation studies discussed in \S\ref{sec: analysis} instead of comparing it with other LLMs.

CCTE encompasses four clinical tasks that, while not specifying the format of model outputs, require a higher level of professional expertise. For this assessment, we employed GPT-4 to score the outputs across four dimensions comprehensively. As shown in Table~\ref{tab: ccte results}, it is evident that even the \ming-1.8B variant outperforms both general and medical-specific large language models, further confirming \ming’s robust potential in clinical settings. More details about CCTE are discussed in Appendix~\ref{appendix: ccte}.




\begin{table*}[tbp]
\centering
\resizebox{\textwidth}{!}{%
\begin{tabular}{cccc|ccccc|cc}
\toprule
\multicolumn{2}{c}{\textbf{MKA}} & \multicolumn{2}{c|}{\textbf{DA}} & \multicolumn{5}{c|}{\textbf{Knowledge-Intensive}} & \multicolumn{2}{c}{\textbf{Alignment-Required}} \\
\textbf{KA \ } & \textbf{\ NA \ } & \small{\textbf{\emph{w/o} $\mathcal{L}_{\mathrm{orth}}$}} & \small{\textbf{\emph{w/} $\mathcal{L}_{\mathrm{orth}}$}} & \textbf{CMMLU} & \textbf{CEval} & \textbf{PLE Pha} & \textbf{PLE TCM} & \textbf{Avg.} & \textbf{CBLUE} & \textbf{CCTE} \\
\midrule
\multicolumn{4}{c|}{Base Model} & 48.45 & 53.66 & 31.46 & 25.63 & 39.80 & 3.76 & 3.36 \\
\midrule
\ding{51} & \ding{55} & \ding{55} & \ding{55} & 46.23 & 41.46 & 33.54 & 33.96 & 38.80 & 51.41 & 4.04 \\
\ding{51} & \ding{51} & \ding{55} & \ding{55} & 50.07 & 45.00 & 36.04 & 33.54 & 41.16 & 57.09 & 4.15 \\
\rowcolor[HTML]{F2F2F2}
\ding{51} & \faCheckSquareO & \ding{55} & \ding{55} & 51.18 & 60.98 & 36.04 & 39.58 & \textbf{46.95} & 8.15 & 3.80 \\
\midrule
\ding{51} & \ding{51} & \ding{55} & \ding{51} & 45.79 & 41.46 & 37.50 & 34.38 & 39.78 & 56.26 & 4.02 \\
\ding{51} & \faCheckSquareO & \ding{51} & \ding{55} & 48.52 & 46.34 & 37.50 & 35.42 & 41.95 & 55.89 & 4.00 \\
\rowcolor[HTML]{F2F2F2}
\ding{51} & \faCheckSquareO & \ding{55} & \ding{51} & 48.74 & 53.66 & 33.33 & 37.08 & 43.20 & \textbf{62.05} & \textbf{4.22} \\
\bottomrule
\end{tabular}%
}
\caption{Ablation Experiments of the \colorbox[HTML]{F2F2F2}{\ming} methods. Note that \faCheckSquareO \ indicates \ming drops \noiseaggregator after MKA fine-tuning stage.}
\label{tab: ablation experiments}
\end{table*}

\subsection{Ablation Experiments}
The results of the ablation experiments are shown in Table~\ref{tab: ablation experiments}. For the methods only fine-tuned with the MKA stage, inference with NA demonstrated minimal improvement in knowledge examination performance. Without the disturbance of the NA, \ming achieved the best knowledge performance but failed to complete the alignment-required tasks. For the two-stage fine-tuning methods~(MKA+DA), further fine-tuned with the NA module or without orthogonal regularization led to a noticeable reduction in performance on knowledge-intensive tasks. The orthogonal regularization can avoid parameter redundancy, facilitating more effective alignment learning while mitigating the loss of reasoning capability.

\begin{table*}[t]
\centering
\resizebox{\textwidth}{!}{%
\begin{tabular}{l|ccccc|cc}
\toprule
\multirow{2}{*}{\textbf{Model}} & \multicolumn{5}{c|}{\textbf{Knowledge-Intensive}} & \multicolumn{2}{c}{\textbf{Alignment-Required}} \\
 & \textbf{CMMLU} & \textbf{CEval} & \textbf{PLE Pha.} & \textbf{PLE TCM.} & \textbf{Avg.} & \textbf{CBLUE} & \textbf{CCTE}  \\
 \midrule
Base Models & 48.45 & 53.66 & 31.46 & 25.63 & 39.80 & 3.76 & 3.36  \\
\midrule
Parallel LoRA & 48.67 & 43.90 & 35.00 & 34.17 & 40.43 & 55.05 & 4.09  \\
\hspace{1em} $\mapsto$  \emph{w/o} LoRA 1 & 51.26 & 51.22 & 34.79 & 35.00 & 43.07 & 11.43 & 3.89  \\
\hspace{1em} $\mapsto$  \emph{w/o} LoRA 2 & 50.52 & 48.78 & 35.63 & 33.75 & 42.17 & 11.87 & 3.88  \\
\midrule
\ming & 50.07 & 45.00 & 36.04 & 33.54 & 41.16 & \textbf{57.09} & \textbf{4.15}  \\
\hspace{1em} $\mapsto$  \emph{w/o} NA & 51.18 & 60.98 & 36.04 & 39.58 & \textbf{46.95} & 8.15 & 3.80  \\
\hspace{1em} $\mapsto$  \emph{w/o} KA & 49.19 & 39.02 & 34.17 & 32.50 & 38.72 & 50.22 & 4.06 \\
\bottomrule
\end{tabular}%
}
\caption{Performance of partial experts with MKA fine-tuning stage. `Parallel LoRA' has two identical LoRAs on FFN modules, which are numbered with LoRA1 and LoRA2 for distinguishment.}
\label{tab: experts capacity}
\end{table*}

\section{Discussion}
\label{sec: analysis}
In this section, we discuss the following research questions~(RQ) of the proposed \ming:
\begin{itemize}
\item \textbf{RQ1:} Why does \noiseaggregator have a negative impact on the task performance?
\item \textbf{RQ2:} Are the roles of \knowledgeaggregator and \noiseaggregator determined by the model architecture?

\item \textbf{RQ3:} How do both two fine-tuning stages improve the model's capabilities?

\item \textbf{RQ4:} Can \ming still learn the knowledge effectively when the scale of fine-tuning corpus becomes smaller?

\item \textbf{RQ5:} How does the effect of \ming compare to other PEFT methods?

\item \textbf{RQ6:} How does \ming perform in other languages?
\end{itemize}

\begin{figure}[t]
    \centering
    \begin{subfigure}{0.5\linewidth}
        \centering
        \includegraphics[width=\linewidth]{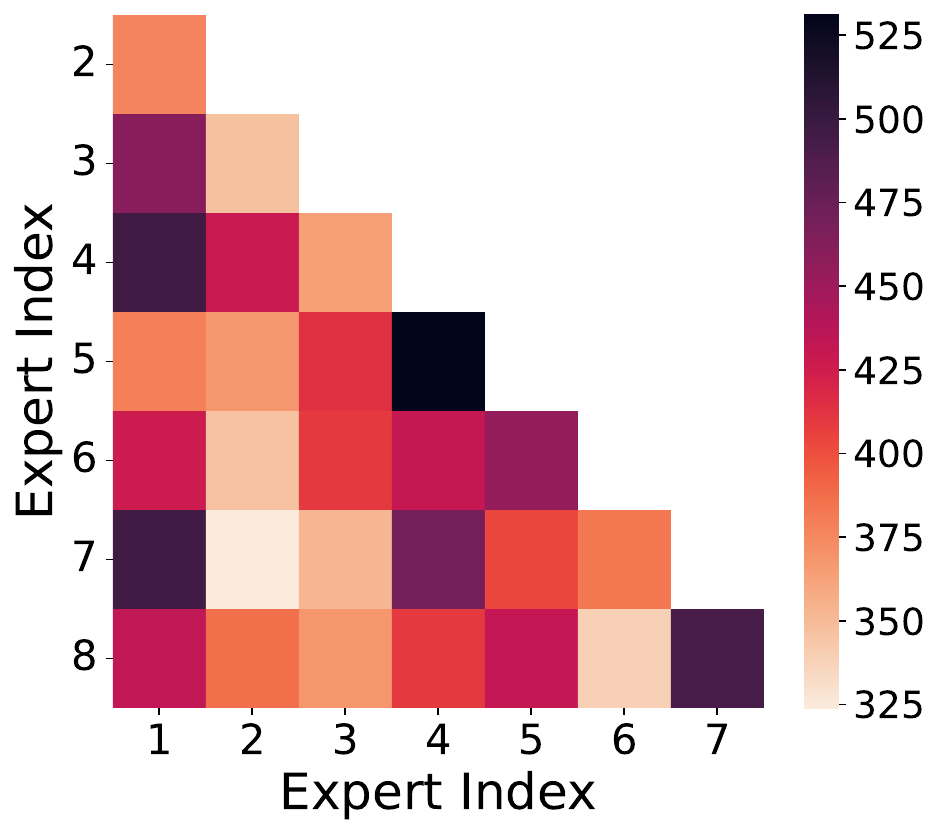}
        \caption{}
        \label{fig: mismatch activation}
    \end{subfigure}%
    \begin{subfigure}{0.5\linewidth}
        \centering
        \includegraphics[width=\linewidth]{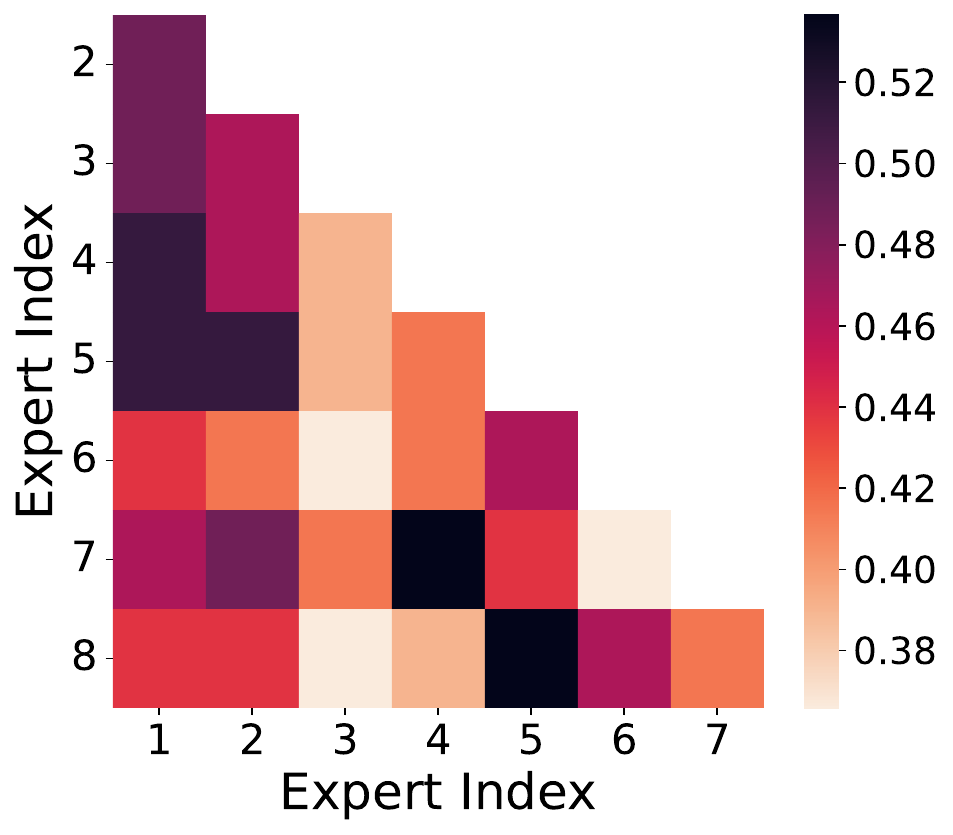}
        \caption{}
        \label{fig: mismatch performance}
    \end{subfigure}
    \caption{Mismatch between expert activation times and performance on CEval. (a) Activation times of each expert combination. (b) Performance of only used each expert combination. The $i$-th raw and $j$-th column indicates the combinations of the $i$-th and $j$-th experts. The accuracy of the vanilla MoLoRA inference is 45.00.
    }
    \label{fig: routing mismatch}
\end{figure}

\begin{figure*}[t]
    \centering
    \begin{subfigure}{0.33\linewidth}
        \centering
        \includegraphics[width=\linewidth]{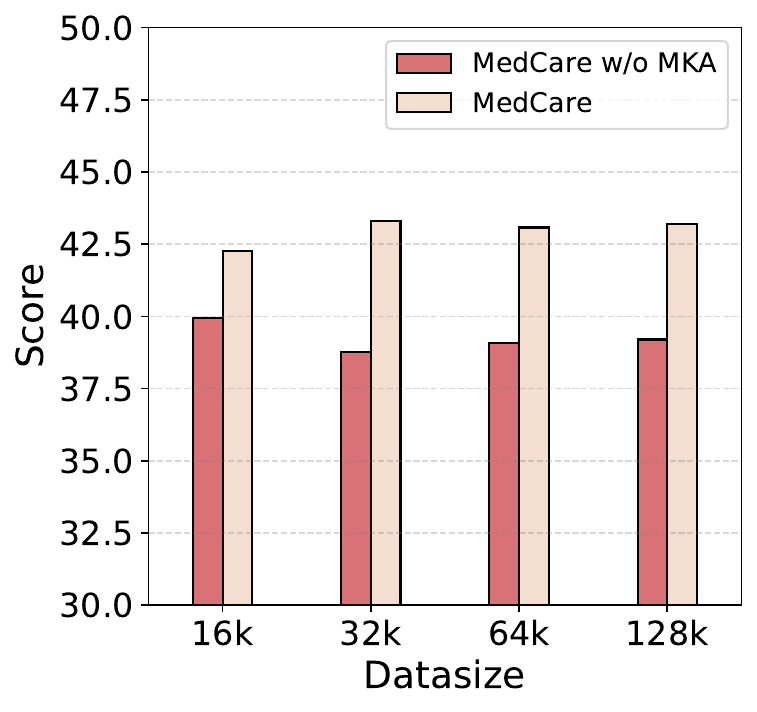}
        \caption{}
        \label{fig: 2stage exam}
    \end{subfigure}%
    \begin{subfigure}{0.32\linewidth}
        \centering
        \includegraphics[width=\linewidth]{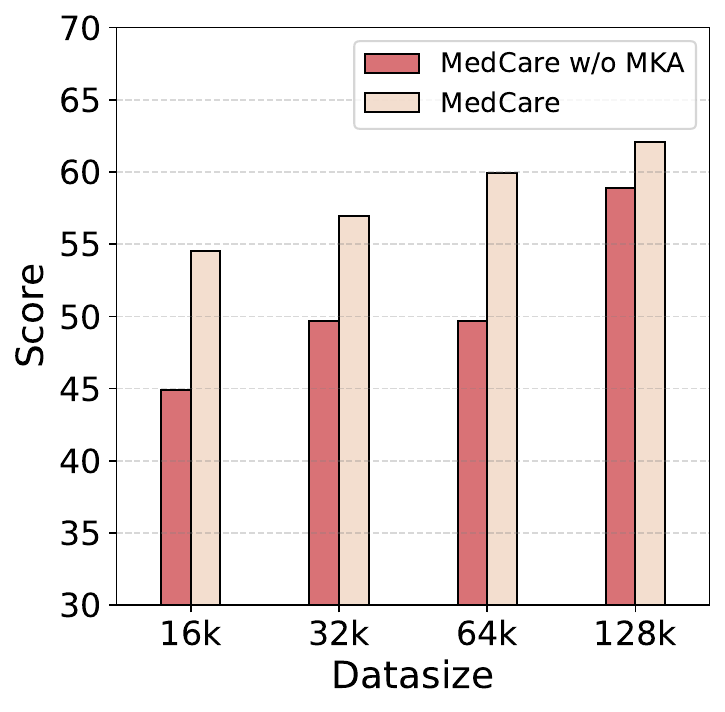}
        \caption{}
        \label{fig: 2stage cblue}
    \end{subfigure}
    \begin{subfigure}{0.32\linewidth}
        \centering
        \includegraphics[width=\linewidth]{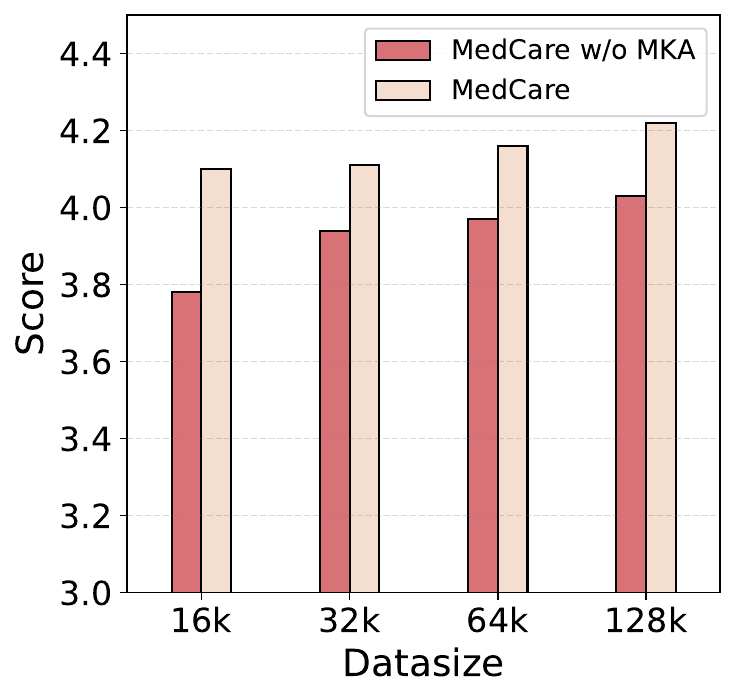}
        \caption{}
        \label{fig: 2stage ccte}
    \end{subfigure}
    \caption{Performance with different sizes of alignment-required datasets for DA fine-tuning stage. (a) The average performance on knowledge-intensive tasks. (b) The average performance on CBLUE. (c) The average performance on CCTE. `\ming \emph{w/o} MKA' indicates the model is fine-tuned with only the second stage. Note that the score of the knowledge examinations is the average of the CMMLU, CEval, PLE Pharamy, and PLE TCM.}
    \label{fig: 2stage dataset}
\end{figure*}

\begin{figure}[t]
    \centering
    \includegraphics[width=0.8\linewidth]{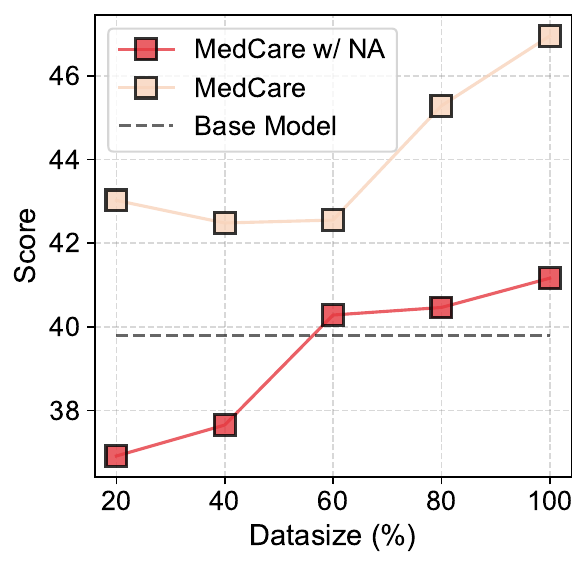}
    \caption{Average performance on the medical knowledge examination with different knowledge aggregation learning data size. `Base Model' indicates the performance of the Qwen1.5-1.8B without fine-tuning.}
    \label{fig: 1stage datasize}
\vspace{-1.0em}
\end{figure}

\paragraph{Response to RQ1: \noiseaggregator interferes with models because of the routing mismatch.}
We investigate the effect of different combinations of MoLoRA experts in \noiseaggregator on the final performance to demonstrate the routing mismatch. We first compute the activation frequencies for each combination of the experts in vanilla MoLoRA inference. Meanwhile, different expert combinations are designated to test the performance of the models. The results in Figure~\ref{fig: routing mismatch} show the mismatch between the experts' activation times and their corresponding performance, which indicates that the router in the MoLoRA module failed to select the experts with optimal performance for each task and even caused the performance drop during inference. By discarding the \noiseaggregator, LLMs can learn knowledge more effectively and surpass the vanilla MoLoRA models with an average of more than 5 points, as shown in Table~\ref{tab: experts capacity}.

\paragraph{Response to RQ2: Yes. Symmetric structure failed to decouple the learning process into the common knowledge and task-specific requirements.
}
To demonstrate this, we tested the models with two parallel LoRA modules to investigate the role of the \noiseaggregator. As shown in Table~\ref{tab: experts capacity}, the parallel LoRA module failed to decouple the learning process due to its symmetric structure. However, it is surprising that removing one of the FFN LoRA modules can improve the model performance on knowledge-intensive tasks. Similar phenomena are also found in previous work~\citep{jiang2024taia}. They found that removing the fine-tuned FFN parameters can fully utilize the general capacity of the base model by producing a more similar hidden state. In the structure of \ming, the \knowledgeaggregator predominantly acquires common knowledge from the fine-tuning corpus, while the \noiseaggregator focuses more on learning alignment formats. 

\paragraph{Response to RQ3: The \ka stage improves the knowledge capacity of the models and the \da stage adapts the models to learn the target formats. }
We explore the respective role of each fine-tuning stage and empirically validate each merit. Figure~\ref{fig: 2stage dataset} shows the results on knowledge examination, CBLUE, and CCTE, respectively. For the first stage, it is obvious that the models with the MKA fine-tuning stage not only achieve better performance on the knowledge examination tasks but also align with the format of the downstream dataset with greater ease.
Although discarding \noiseaggregator after the first stage of fine-tuning reduces the alignment performance of the model, the target format can be learned faster by retraining the model with the second-stage medical adaptation. 
As more second-stage aligned data is added, the model's alignment capability significantly improves without compromising its knowledge capacity. 
This suggests that the second stage of the fine-tuning process primarily helps the model align with the format, rather than learning new knowledge.

\begin{table*}[tbp]
\centering
\resizebox{\textwidth}{!}{%
\begin{tabular}{l|ccccc|cccc}
\toprule
 & \multicolumn{5}{c|}{\textbf{Knowledge-Intensive}} & \multicolumn{4}{c}{\textbf{Cross-Lingual Generalization}} \\
\multirow{-2}{*}{\textbf{Model}} & \textbf{CMMLU} & \textbf{CEval} & \textbf{PLE Pha} & \textbf{PLE TCM} & \textbf{Avg.} & \textbf{MedQA} & \textbf{MedMCQA} & \textbf{MMLU} & \textbf{Avg.} \\
\midrule
Qwen1.5-7B & \textbf{66.77} & 68.29 & 32.29 & 29.17 & 49.13 & 40.46 & 45.33 & 61.43 & 49.07 \\
Huatuo-II-7B & 59.08 & 62.40 & 47.70 & 47.50 & 54.17 & 40.69 & 44.75 & 44.33 & 43.26 \\
\midrule
LoRA~\citep{hu2021lora} & 65.21 & 60.98 & 46.67 & 50.83 & 55.92 & 40.53 & 44.94 & 61.24 & 48.90 \\
LoRA+~\citep{DBLP:conf/icml/HayouG024} & 63.29 & 68.29 & 45.63 & 45.63 & 55.71 & 38.65 & 45.21 & 61.69 & 48.52 \\
MoRA~\citep{jiang2024mora} & 65.07 & 63.41 & 49.17 & 47.92 & 56.39 & 38.65 & 45.66 & 61.11 & 48.47 \\
DoRA~\citep{DBLP:conf/icml/LiuWY0WCC24} & 65.14 & \textbf{70.73} & 50.00 & 50.83 & 59.18 & 39.75 & 46.02 & 60.60 & 48.79 \\
\rowcolor[HTML]{F2F2F2}
\ming & 66.25 & \textbf{70.73} & \textbf{50.83} & \textbf{53.33} & \textbf{60.29} & \textbf{40.96} & \textbf{46.98} & \textbf{62.27} & \textbf{50.07} \\
\bottomrule
\end{tabular}%
}
\caption{Comparison with other parameter efficient fine-tuning~(PEFT) methods on knowledge-intensive tasks and cross-lingual generalization. The results of PEFT methods are fine-tuned on Qwen1.5-7B. The datasets in cross-lingual generalization are all in English. Note that the `MedQA' indicates the US subset.}
\label{tab: peft comparison}
\end{table*}

\paragraph{Response to RQ4: Yes. Even with limited data, \ming can enhance its medical knowledge proficiency.}
To show the effective knowledge learning ability of \ming, we compare the performance of models with varying scales of the first-stage fine-tuning corpus, and present the results in Figure~\ref{fig: 1stage datasize}. The vanilla MoLoRA suffers from performance degradation when the size of the training-corpus is smaller than 50\%, while \ming utilizes the knowledge in the corpus among all sizes of the data and surpasses the performance of the base model consistently, demonstrating that \ming has better knowledge utilization ability.

\paragraph{Response to RQ5: \ming outperforms the commonly used PEFT methods on knowledge-intensive tasks.} 
We choose four PEFT methods to show the effectiveness of \ming, including LoRA~\citep{hu2021lora}, LoRA+~\citep{DBLP:conf/icml/HayouG024}, MoRA~\citep{jiang2024mora}, and DoRA~\citep{DBLP:conf/icml/LiuWY0WCC24}. It is observed in Table~\ref{tab: peft comparison} that most of the PEFT methods suffer from performance drops on the medical parts of the general benchmarks CMMLU and CEval. This shows that domain adaptation will destroy the generalization ability of LLM and reduce its ability in scenarios outside the distribution of training data. The more effective methods, like DoRA and \ming, can not only improve the domain capacity but also keep the generalization ability of the LLMs simultaneously, where \ming demonstrates more superiority than the DoRA method.

\paragraph{Response to RQ6: \ming shows strong cross-lingual generalization ability with only fine-tuned with monolingual data.}
We choose three English medical datasets, MedQA~\citep{jin2021disease}, MedMCQA~\citep{pmlr-v174-pal22a}, and the medical part of MMLU~\citep{hendryckstest2021}, to validate the cross-lingual capacity of the LLMs. The results in Table~\ref{tab: peft comparison} show that all the other PEFT methods with only Chinese fine-tuning data achieve worse performance on English medical benchmarks when compared to the base model Qwen1.5-7B. However, with only Chinese fine-tuning data, \ming can still improve the model's performance on the English dataset with 1 point, showing strong cross-lingual ability and applicability.

\section{Conclusions}
In this paper, we first categorize medical tasks into knowledge-intensive tasks and alignment-required tasks to reinforce the practical usability requirements for medical LLMs. Then, we propose a two-stage fine-tuning framework to adapt LLMs to medical domains and mitigate knowledge performance degradation and propose \ming. The experiment results show that \ming can achieve promising performance on over 20 knowledge-intensive and alignment-required tasks.

\section*{Limitations}
In this paper, we propose a two-stage fine-tuning framework that mitigates the damage and loss of pre-trained knowledge in large language models during downstream fine-tuning for medical tasks. Despite its effectiveness, our approach still exhibits limitations. Through knowledge-aggregative learning, our method enables large language models to more effectively assimilate the knowledge of the fine-tuning corpus, thereby enhancing the knowledge capabilities of the final model beyond the baseline. However, the alignment fine-tuning in the second stage still adversely affects the model's knowledge capabilities. Furthermore, our method does not yet allow for the decoupling of knowledge and format learning directly, but instead requires two-stage training. Addressing these two issues will be the focus of our future work.

\section*{Ethic Considerations}
In this article, we introduce a medical large language model, designated as \ming, which incorporates several ethical considerations crucial for its deployment and usage in sensitive settings:

\paragraph{Performance vs. Potential Risks} While \ming demonstrates enhancements over previous general and specialized medical models in knowledge reasoning and performance on downstream tasks, it's important to acknowledge the inherent limitations of large language models. Notably, these models can exhibit "hallucinations" or generate misleading information. Additionally, the training datasets might harbor undiscovered biases, which could inadvertently influence the model's outputs. Given these concerns, we emphasize that \ming is not suitable for providing medical advice or for use in direct clinical applications. 

\paragraph{Data Ethics and Privacy Compliance} The datasets employed in this study, including the knowledge testing data and the CBLUE dataset, are all publicly available and open-source and thus do not pose ethical dilemmas concerning their usage. However, the clinical data from CCTE used in training and testing, which involves report generation, image analysis, and discharge instructions, originates from hospital inpatient records. We have taken stringent measures to ensure the privacy and confidentiality of this information. All personal identifiers have been removed to maintain anonymity, ensuring no individual can be recognized from the data used. During the data collection, patients signed informed consent forms and were fully aware of the data usage methods described in this paper. Additionally, the usage of this data has been reviewed and approved by the corresponding hospital ethics committees. The specific approval numbers will be provided after the end of the review. This ensures that the data usage in this paper fully complies with ethical standards and privacy protection regulations.

\section*{Acknowledgements}
This work is supported by the National Key R\&D Program of China (No. 2022ZD0162101), the National Natural Science Foundation of China (No. 62106140), and STCSM (No. 21511101100, No. 22DZ2229005)

\bibliography{custom}

\begin{thebibliography}{70}
\providecommand{\natexlab}[1]{#1}

\bibitem[{AI(2024)}]{metaIntroducingMeta}
Meta AI. 2024.
\newblock {I}ntroducing {M}eta {L}lama 3: {T}he most capable openly available {L}{L}{M} to date --- ai.meta.com.
\newblock \url{https://ai.meta.com/blog/meta-llama-3/}.
\newblock [Accessed 05-06-2024].

\bibitem[{Bai et~al.(2023)Bai, Bai, Chu, Cui, Dang, Deng, Fan, Ge, Han, Huang et~al.}]{bai2023qwen}
Jinze Bai, Shuai Bai, Yunfei Chu, Zeyu Cui, Kai Dang, Xiaodong Deng, Yang Fan, Wenbin Ge, Yu~Han, Fei Huang, et~al. 2023.
\newblock Qwen technical report.
\newblock \emph{arXiv preprint arXiv:2309.16609}.

\bibitem[{Cai et~al.(2023)Cai, Chen, Guo, Wang, Li, Liu, Zheng, Liu, and Chen}]{cai2023regemr}
Jie Cai, Shenglin Chen, Siyun Guo, Suidong Wang, Lintong Li, Xiaotong Liu, Keming Zheng, Yudong Liu, and Shiling Chen. 2023.
\newblock Regemr: a natural language processing system to automatically identify premature ovarian decline from chinese electronic medical records.
\newblock \emph{BMC Medical Informatics and Decision Making}, 23(1):126.

\bibitem[{Chen et~al.(2023{\natexlab{a}})Chen, Wang, Gao, Jiang, Chen, Zhang, Song, Xie, Kong, Li et~al.}]{chen2023huatuogpt}
Junying Chen, Xidong Wang, Anningzhe Gao, Feng Jiang, Shunian Chen, Hongbo Zhang, Dingjie Song, Wenya Xie, Chuyi Kong, Jianquan Li, et~al. 2023{\natexlab{a}}.
\newblock Huatuogpt-ii, one-stage training for medical adaption of llms.
\newblock \emph{arXiv preprint arXiv:2311.09774}.

\bibitem[{Chen et~al.(2023{\natexlab{b}})Chen, Wang, Xing, Xu, Fang, Wang, Li, Wu, Liu, Xu et~al.}]{chen2023bianque}
Yirong Chen, Zhenyu Wang, Xiaofen Xing, Zhipei Xu, Kai Fang, Junhong Wang, Sihang Li, Jieling Wu, Qi~Liu, Xiangmin Xu, et~al. 2023{\natexlab{b}}.
\newblock Bianque: Balancing the questioning and suggestion ability of health llms with multi-turn health conversations polished by chatgpt.
\newblock \emph{arXiv preprint arXiv:2310.15896}.

\bibitem[{Chowdhery et~al.(2023)Chowdhery, Narang, Devlin, Bosma, Mishra, Roberts, Barham, Chung, Sutton, Gehrmann et~al.}]{chowdhery2023palm}
Aakanksha Chowdhery, Sharan Narang, Jacob Devlin, Maarten Bosma, Gaurav Mishra, Adam Roberts, Paul Barham, Hyung~Won Chung, Charles Sutton, Sebastian Gehrmann, et~al. 2023.
\newblock Palm: Scaling language modeling with pathways.
\newblock \emph{Journal of Machine Learning Research}, 24(240):1--113.

\bibitem[{Chung et~al.(2022)Chung, Hou, Longpre, Zoph, Tay, Fedus, Li, Wang, Dehghani, Brahma et~al.}]{chung2022scaling}
Hyung~Won Chung, Le~Hou, Shayne Longpre, Barret Zoph, Yi~Tay, William Fedus, Yunxuan Li, Xuezhi Wang, Mostafa Dehghani, Siddhartha Brahma, et~al. 2022.
\newblock Scaling instruction-finetuned language models.
\newblock \emph{arXiv preprint arXiv:2210.11416}.

\bibitem[{Dai et~al.(2024)Dai, Deng, Zhao, Xu, Gao, Chen, Li, Zeng, Yu, Wu et~al.}]{dai2024deepseekmoe}
Damai Dai, Chengqi Deng, Chenggang Zhao, RX~Xu, Huazuo Gao, Deli Chen, Jiashi Li, Wangding Zeng, Xingkai Yu, Y~Wu, et~al. 2024.
\newblock Deepseekmoe: Towards ultimate expert specialization in mixture-of-experts language models.
\newblock \emph{arXiv preprint arXiv:2401.06066}.

\bibitem[{Deng et~al.(2024)Deng, Zhang, He, Chen, Shi, Xu, Fu, Zhang, Wang, Zhou et~al.}]{deng2024k2}
Cheng Deng, Tianhang Zhang, Zhongmou He, Qiyuan Chen, Yuanyuan Shi, Yi~Xu, Luoyi Fu, Weinan Zhang, Xinbing Wang, Chenghu Zhou, et~al. 2024.
\newblock K2: A foundation language model for geoscience knowledge understanding and utilization.
\newblock In \emph{Proceedings of the 17th ACM International Conference on Web Search and Data Mining}, pages 161--170.

\bibitem[{Dou et~al.(2023)Dou, Zhou, Liu, Gao, Zhao, Shen, Zhou, Xi, Wang, Fan et~al.}]{dou2023loramoe}
Shihan Dou, Enyu Zhou, Yan Liu, Songyang Gao, Jun Zhao, Wei Shen, Yuhao Zhou, Zhiheng Xi, Xiao Wang, Xiaoran Fan, et~al. 2023.
\newblock Loramoe: Revolutionizing mixture of experts for maintaining world knowledge in language model alignment.
\newblock \emph{arXiv preprint arXiv:2312.09979}.

\bibitem[{Elfwing et~al.(2018)Elfwing, Uchibe, and Doya}]{elfwing2018sigmoid}
Stefan Elfwing, Eiji Uchibe, and Kenji Doya. 2018.
\newblock Sigmoid-weighted linear units for neural network function approximation in reinforcement learning.
\newblock \emph{Neural networks}, 107:3--11.

\bibitem[{Feng et~al.(2024)Feng, Hao, Zhang, Han, and Wang}]{feng-etal-2024-mixture-loras}
Wenfeng Feng, Chuzhan Hao, Yuewei Zhang, Yu~Han, and Hao Wang. 2024.
\newblock \href {https://aclanthology.org/2024.lrec-main.994} {Mixture-of-{L}o{RA}s: An efficient multitask tuning method for large language models}.
\newblock In \emph{Proceedings of the 2024 Joint International Conference on Computational Linguistics, Language Resources and Evaluation (LREC-COLING 2024)}, pages 11371--11380, Torino, Italia. ELRA and ICCL.

\bibitem[{Gekhman et~al.(2024)Gekhman, Yona, Aharoni, Eyal, Feder, Reichart, and Herzig}]{gekhman2024does}
Zorik Gekhman, Gal Yona, Roee Aharoni, Matan Eyal, Amir Feder, Roi Reichart, and Jonathan Herzig. 2024.
\newblock Does fine-tuning llms on new knowledge encourage hallucinations?
\newblock \emph{arXiv preprint arXiv:2405.05904}.

\bibitem[{Geva et~al.(2020)Geva, Schuster, Berant, and Levy}]{geva2020transformer}
Mor Geva, Roei Schuster, Jonathan Berant, and Omer Levy. 2020.
\newblock Transformer feed-forward layers are key-value memories.
\newblock \emph{arXiv preprint arXiv:2012.14913}.

\bibitem[{Gou et~al.(2023)Gou, Liu, Chen, Hong, Xu, Li, Yeung, Kwok, and Zhang}]{gou2023mixture}
Yunhao Gou, Zhili Liu, Kai Chen, Lanqing Hong, Hang Xu, Aoxue Li, Dit-Yan Yeung, James~T Kwok, and Yu~Zhang. 2023.
\newblock Mixture of cluster-conditional lora experts for vision-language instruction tuning.
\newblock \emph{arXiv preprint arXiv:2312.12379}.

\bibitem[{Hayou et~al.(2024)Hayou, Ghosh, and Yu}]{DBLP:conf/icml/HayouG024}
Soufiane Hayou, Nikhil Ghosh, and Bin Yu. 2024.
\newblock \href {https://openreview.net/forum?id=NEv8YqBROO} {Lora+: Efficient low rank adaptation of large models}.
\newblock In \emph{Forty-first International Conference on Machine Learning, {ICML} 2024, Vienna, Austria, July 21-27, 2024}. OpenReview.net.

\bibitem[{Hendrycks et~al.(2021)Hendrycks, Burns, Basart, Zou, Mazeika, Song, and Steinhardt}]{hendryckstest2021}
Dan Hendrycks, Collin Burns, Steven Basart, Andy Zou, Mantas Mazeika, Dawn Song, and Jacob Steinhardt. 2021.
\newblock Measuring massive multitask language understanding.
\newblock \emph{Proceedings of the International Conference on Learning Representations (ICLR)}.

\bibitem[{Hongying et~al.(2021)Hongying, Wenxin, Kunli, Yajuan, Baobao, and Zhifang}]{hongying2021building}
Zan Hongying, Li~Wenxin, Zhang Kunli, Ye~Yajuan, Chang Baobao, and Sui Zhifang. 2021.
\newblock Building a pediatric medical corpus: Word segmentation and named entity annotation.
\newblock In \emph{Chinese Lexical Semantics: 21st Workshop, CLSW 2020, Hong Kong, China, May 28--30, 2020, Revised Selected Papers 21}, pages 652--664. Springer.

\bibitem[{Houlsby et~al.(2019)Houlsby, Giurgiu, Jastrzebski, Morrone, De~Laroussilhe, Gesmundo, Attariyan, and Gelly}]{houlsby2019parameter}
Neil Houlsby, Andrei Giurgiu, Stanislaw Jastrzebski, Bruna Morrone, Quentin De~Laroussilhe, Andrea Gesmundo, Mona Attariyan, and Sylvain Gelly. 2019.
\newblock Parameter-efficient transfer learning for nlp.
\newblock In \emph{International conference on machine learning}, pages 2790--2799. PMLR.

\bibitem[{Hu et~al.(2021)Hu, Shen, Wallis, Allen-Zhu, Li, Wang, Wang, and Chen}]{hu2021lora}
Edward~J Hu, Yelong Shen, Phillip Wallis, Zeyuan Allen-Zhu, Yuanzhi Li, Shean Wang, Lu~Wang, and Weizhu Chen. 2021.
\newblock Lora: Low-rank adaptation of large language models.
\newblock \emph{arXiv preprint arXiv:2106.09685}.

\bibitem[{Hu et~al.(2022)Hu, yelong shen, Wallis, Allen-Zhu, Li, Wang, Wang, and Chen}]{hu2022lora}
Edward~J Hu, yelong shen, Phillip Wallis, Zeyuan Allen-Zhu, Yuanzhi Li, Shean Wang, Lu~Wang, and Weizhu Chen. 2022.
\newblock \href {https://openreview.net/forum?id=nZeVKeeFYf9} {Lo{RA}: Low-rank adaptation of large language models}.
\newblock In \emph{International Conference on Learning Representations}.

\bibitem[{Huang et~al.(2023)Huang, Bai, Zhu, Zhang, Zhang, Su, Liu, Lv, Zhang, Lei, Fu, Sun, and He}]{huang2023ceval}
Yuzhen Huang, Yuzhuo Bai, Zhihao Zhu, Junlei Zhang, Jinghan Zhang, Tangjun Su, Junteng Liu, Chuancheng Lv, Yikai Zhang, Jiayi Lei, Yao Fu, Maosong Sun, and Junxian He. 2023.
\newblock C-eval: A multi-level multi-discipline chinese evaluation suite for foundation models.
\newblock In \emph{Advances in Neural Information Processing Systems}.

\bibitem[{Jiang et~al.(2024{\natexlab{a}})Jiang, Liao, Zhang, Wang, and Wang}]{jiang2024taia}
Shuyang Jiang, Yusheng Liao, Ya~Zhang, Yu~Wang, and Yanfeng Wang. 2024{\natexlab{a}}.
\newblock Taia: Large language models are out-of-distribution data learners.
\newblock \emph{arXiv preprint arXiv:2405.20192}.

\bibitem[{Jiang et~al.(2024{\natexlab{b}})Jiang, Huang, Luo, Zhang, Huang, Wei, Deng, Sun, Zhang, Wang et~al.}]{jiang2024mora}
Ting Jiang, Shaohan Huang, Shengyue Luo, Zihan Zhang, Haizhen Huang, Furu Wei, Weiwei Deng, Feng Sun, Qi~Zhang, Deqing Wang, et~al. 2024{\natexlab{b}}.
\newblock Mora: High-rank updating for parameter-efficient fine-tuning.
\newblock \emph{arXiv preprint arXiv:2405.12130}.

\bibitem[{Jin et~al.(2021)Jin, Pan, Oufattole, Weng, Fang, and Szolovits}]{jin2021disease}
Di~Jin, Eileen Pan, Nassim Oufattole, Wei-Hung Weng, Hanyi Fang, and Peter Szolovits. 2021.
\newblock What disease does this patient have? a large-scale open domain question answering dataset from medical exams.
\newblock \emph{Applied Sciences}, 11(14):6421.

\bibitem[{Labrak et~al.(2024)Labrak, Bazoge, Morin, Gourraud, Rouvier, and Dufour}]{labrak2024biomistral}
Yanis Labrak, Adrien Bazoge, Emmanuel Morin, Pierre-Antoine Gourraud, Mickael Rouvier, and Richard Dufour. 2024.
\newblock Biomistral: A collection of open-source pretrained large language models for medical domains.
\newblock \emph{arXiv preprint arXiv:2402.10373}.

\bibitem[{Lester et~al.(2021)Lester, Al-Rfou, and Constant}]{lester-etal-2021-power}
Brian Lester, Rami Al-Rfou, and Noah Constant. 2021.
\newblock \href {https://doi.org/10.18653/v1/2021.emnlp-main.243} {The power of scale for parameter-efficient prompt tuning}.
\newblock In \emph{Proceedings of the 2021 Conference on Empirical Methods in Natural Language Processing}, pages 3045--3059, Online and Punta Cana, Dominican Republic. Association for Computational Linguistics.

\bibitem[{Li et~al.(2024)Li, Ma, Wang, Cheng, Duan, Zuo, Yang, and Tang}]{li2024mixlora}
Dengchun Li, Yingzi Ma, Naizheng Wang, Zhiyuan Cheng, Lei Duan, Jie Zuo, Cal Yang, and Mingjie Tang. 2024.
\newblock Mixlora: Enhancing large language models fine-tuning with lora based mixture of experts.
\newblock \emph{arXiv preprint arXiv:2404.15159}.

\bibitem[{Li et~al.(2023{\natexlab{a}})Li, Zhang, Koto, Yang, Zhao, Gong, Duan, and Baldwin}]{li2023cmmlu}
Haonan Li, Yixuan Zhang, Fajri Koto, Yifei Yang, Hai Zhao, Yeyun Gong, Nan Duan, and Timothy Baldwin. 2023{\natexlab{a}}.
\newblock Cmmlu: Measuring massive multitask language understanding in chinese.
\newblock \emph{arXiv preprint arXiv:2306.09212}.

\bibitem[{Li and Liang(2021)}]{li-liang-2021-prefix}
Xiang~Lisa Li and Percy Liang. 2021.
\newblock \href {https://doi.org/10.18653/v1/2021.acl-long.353} {Prefix-tuning: Optimizing continuous prompts for generation}.
\newblock In \emph{Proceedings of the 59th Annual Meeting of the Association for Computational Linguistics and the 11th International Joint Conference on Natural Language Processing (Volume 1: Long Papers)}, pages 4582--4597, Online. Association for Computational Linguistics.

\bibitem[{Li et~al.(2023{\natexlab{b}})Li, Li, Zhang, Dan, Jiang, and Zhang}]{li2023chatdoctor}
Yunxiang Li, Zihan Li, Kai Zhang, Ruilong Dan, Steve Jiang, and You Zhang. 2023{\natexlab{b}}.
\newblock Chatdoctor: A medical chat model fine-tuned on a large language model meta-ai (llama) using medical domain knowledge.
\newblock \emph{Cureus}, 15(6).

\bibitem[{Lin et~al.(2023)Lin, Tan, Lin, Zheng, Pi, Zhang, Diao, Wang, Zhao, Yao et~al.}]{lin2023speciality}
Yong Lin, Lu~Tan, Hangyu Lin, Zeming Zheng, Renjie Pi, Jipeng Zhang, Shizhe Diao, Haoxiang Wang, Han Zhao, Yuan Yao, et~al. 2023.
\newblock Speciality vs generality: An empirical study on catastrophic forgetting in fine-tuning foundation models.
\newblock \emph{arXiv preprint arXiv:2309.06256}.

\bibitem[{Lin et~al.(2020)Lin, Madotto, and Fung}]{lin-etal-2020-exploring}
Zhaojiang Lin, Andrea Madotto, and Pascale Fung. 2020.
\newblock \href {https://doi.org/10.18653/v1/2020.findings-emnlp.41} {Exploring versatile generative language model via parameter-efficient transfer learning}.
\newblock In \emph{Findings of the Association for Computational Linguistics: EMNLP 2020}, pages 441--459, Online. Association for Computational Linguistics.

\bibitem[{Liu et~al.(2023{\natexlab{a}})Liu, Liao, Meng, and Wang}]{LAWGPT-zh}
Hongcheng Liu, Yusheng Liao, Yutong Meng, and Yuhao Wang. 2023{\natexlab{a}}.
\newblock Lawgpt: Chinese law large language models.
\newblock \url{https://github.com/LiuHC0428/LAW_GPT}.

\bibitem[{Liu et~al.(2024{\natexlab{a}})Liu, Zhou, Hua, Chong, Tian, Liu, Wang, You, Guo, Zhu et~al.}]{liu2024benchmarking}
Junling Liu, Peilin Zhou, Yining Hua, Dading Chong, Zhongyu Tian, Andrew Liu, Helin Wang, Chenyu You, Zhenhua Guo, Lei Zhu, et~al. 2024{\natexlab{a}}.
\newblock Benchmarking large language models on cmexam-a comprehensive chinese medical exam dataset.
\newblock \emph{Advances in Neural Information Processing Systems}, 36.

\bibitem[{Liu et~al.(2023{\natexlab{b}})Liu, Wu, Zhao, Zhu, Xu, Tian, and Zheng}]{liu2023moelora}
Qidong Liu, Xian Wu, Xiangyu Zhao, Yuanshao Zhu, Derong Xu, Feng Tian, and Yefeng Zheng. 2023{\natexlab{b}}.
\newblock Moelora: An moe-based parameter efficient fine-tuning method for multi-task medical applications.
\newblock \emph{arXiv preprint arXiv:2310.18339}.

\bibitem[{Liu et~al.(2024{\natexlab{b}})Liu, Wang, Yin, Molchanov, Wang, Cheng, and Chen}]{DBLP:conf/icml/LiuWY0WCC24}
Shih{-}Yang Liu, Chien{-}Yi Wang, Hongxu Yin, Pavlo Molchanov, Yu{-}Chiang~Frank Wang, Kwang{-}Ting Cheng, and Min{-}Hung Chen. 2024{\natexlab{b}}.
\newblock \href {https://openreview.net/forum?id=3d5CIRG1n2} {Dora: Weight-decomposed low-rank adaptation}.
\newblock In \emph{Forty-first International Conference on Machine Learning, {ICML} 2024, Vienna, Austria, July 21-27, 2024}. OpenReview.net.

\bibitem[{Liu et~al.(2024{\natexlab{c}})Liu, Wang, Yin, Molchanov, Wang, Cheng, and Chen}]{liu2024dora}
Shih-Yang Liu, Chien-Yi Wang, Hongxu Yin, Pavlo Molchanov, Yu-Chiang~Frank Wang, Kwang-Ting Cheng, and Min-Hung Chen. 2024{\natexlab{c}}.
\newblock Dora: Weight-decomposed low-rank adaptation.
\newblock \emph{arXiv preprint arXiv:2402.09353}.

\bibitem[{Liu et~al.(2022{\natexlab{a}})Liu, Tang, Cheng, Li, Zheng, and Liang}]{liu2022meddg}
Wenge Liu, Jianheng Tang, Yi~Cheng, Wenjie Li, Yefeng Zheng, and Xiaodan Liang. 2022{\natexlab{a}}.
\newblock Meddg: an entity-centric medical consultation dataset for entity-aware medical dialogue generation.
\newblock In \emph{CCF International Conference on Natural Language Processing and Chinese Computing}, pages 447--459. Springer.

\bibitem[{Liu et~al.(2022{\natexlab{b}})Liu, Ji, Fu, Tam, Du, Yang, and Tang}]{liu-etal-2022-p}
Xiao Liu, Kaixuan Ji, Yicheng Fu, Weng Tam, Zhengxiao Du, Zhilin Yang, and Jie Tang. 2022{\natexlab{b}}.
\newblock \href {https://doi.org/10.18653/v1/2022.acl-short.8} {{P}-tuning: Prompt tuning can be comparable to fine-tuning across scales and tasks}.
\newblock In \emph{Proceedings of the 60th Annual Meeting of the Association for Computational Linguistics (Volume 2: Short Papers)}, pages 61--68, Dublin, Ireland. Association for Computational Linguistics.

\bibitem[{Liu et~al.(2023{\natexlab{c}})Liu, Zheng, Du, Ding, Qian, Yang, and Tang}]{LIU2023}
Xiao Liu, Yanan Zheng, Zhengxiao Du, Ming Ding, Yujie Qian, Zhilin Yang, and Jie Tang. 2023{\natexlab{c}}.
\newblock \href {https://doi.org/10.1016/j.aiopen.2023.08.012} {Gpt understands, too}.
\newblock \emph{AI Open}.

\bibitem[{Luo et~al.(2024)Luo, Lei, Lei, Liu, He, Zhao, and Liu}]{luo2024moelora}
Tongxu Luo, Jiahe Lei, Fangyu Lei, Weihao Liu, Shizhu He, Jun Zhao, and Kang Liu. 2024.
\newblock Moelora: Contrastive learning guided mixture of experts on parameter-efficient fine-tuning for large language models.
\newblock \emph{arXiv preprint arXiv:2402.12851}.

\bibitem[{OpenAI(2022)}]{elmohamed}
OpenAI. 2022.
\newblock Chatgpt: Optimizing language models for dialogue.
\newblock Website.
\newblock \url{https://openai.com/blog/chatgpt}.

\bibitem[{Pal et~al.(2022)Pal, Umapathi, and Sankarasubbu}]{pmlr-v174-pal22a}
Ankit Pal, Logesh~Kumar Umapathi, and Malaikannan Sankarasubbu. 2022.
\newblock \href {https://proceedings.mlr.press/v174/pal22a.html} {Medmcqa: A large-scale multi-subject multi-choice dataset for medical domain question answering}.
\newblock In \emph{Proceedings of the Conference on Health, Inference, and Learning}, volume 174 of \emph{Proceedings of Machine Learning Research}, pages 248--260. PMLR.

\bibitem[{Pfeiffer et~al.(2021)Pfeiffer, Kamath, R{\"u}ckl{\'e}, Cho, and Gurevych}]{pfeiffer-etal-2021-adapterfusion}
Jonas Pfeiffer, Aishwarya Kamath, Andreas R{\"u}ckl{\'e}, Kyunghyun Cho, and Iryna Gurevych. 2021.
\newblock \href {https://doi.org/10.18653/v1/2021.eacl-main.39} {{A}dapter{F}usion: Non-destructive task composition for transfer learning}.
\newblock In \emph{Proceedings of the 16th Conference of the European Chapter of the Association for Computational Linguistics: Main Volume}, pages 487--503, Online. Association for Computational Linguistics.

\bibitem[{Qiu et~al.(2024)Qiu, Wu, Zhang, Lin, Wang, Zhang, Wang, and Xie}]{qiu2024towards}
Pengcheng Qiu, Chaoyi Wu, Xiaoman Zhang, Weixiong Lin, Haicheng Wang, Ya~Zhang, Yanfeng Wang, and Weidi Xie. 2024.
\newblock Towards building multilingual language model for medicine.
\newblock \emph{arXiv preprint arXiv:2402.13963}.

\bibitem[{Rebuffi et~al.(2017)Rebuffi, Bilen, and Vedaldi}]{rebuffi2017learning}
Sylvestre-Alvise Rebuffi, Hakan Bilen, and Andrea Vedaldi. 2017.
\newblock Learning multiple visual domains with residual adapters.
\newblock \emph{Advances in neural information processing systems}, 30.

\bibitem[{Sanh et~al.(2021)Sanh, Webson, Raffel, Bach, Sutawika, Alyafeai, Chaffin, Stiegler, Scao, Raja et~al.}]{sanh2021multitask}
Victor Sanh, Albert Webson, Colin Raffel, Stephen~H Bach, Lintang Sutawika, Zaid Alyafeai, Antoine Chaffin, Arnaud Stiegler, Teven~Le Scao, Arun Raja, et~al. 2021.
\newblock Multitask prompted training enables zero-shot task generalization.
\newblock \emph{arXiv preprint arXiv:2110.08207}.

\bibitem[{Singhal et~al.(2022)Singhal, Azizi, Tu, Mahdavi, Wei, Chung, Scales, Tanwani, Cole-Lewis, Pfohl et~al.}]{singhal2022large}
Karan Singhal, Shekoofeh Azizi, Tao Tu, S~Sara Mahdavi, Jason Wei, Hyung~Won Chung, Nathan Scales, Ajay Tanwani, Heather Cole-Lewis, Stephen Pfohl, et~al. 2022.
\newblock Large language models encode clinical knowledge.
\newblock \emph{arXiv preprint arXiv:2212.13138}.

\bibitem[{Singhal et~al.(2023{\natexlab{a}})Singhal, Azizi, Tu, Mahdavi, Wei, Chung, Scales, Tanwani, Cole-Lewis, Pfohl et~al.}]{singhal2023large}
Karan Singhal, Shekoofeh Azizi, Tao Tu, S~Sara Mahdavi, Jason Wei, Hyung~Won Chung, Nathan Scales, Ajay Tanwani, Heather Cole-Lewis, Stephen Pfohl, et~al. 2023{\natexlab{a}}.
\newblock Large language models encode clinical knowledge.
\newblock \emph{Nature}, 620(7972):172--180.

\bibitem[{Singhal et~al.(2023{\natexlab{b}})Singhal, Tu, Gottweis, Sayres, Wulczyn, Hou, Clark, Pfohl, Cole-Lewis, Neal et~al.}]{singhal2023towards}
Karan Singhal, Tao Tu, Juraj Gottweis, Rory Sayres, Ellery Wulczyn, Le~Hou, Kevin Clark, Stephen Pfohl, Heather Cole-Lewis, Darlene Neal, et~al. 2023{\natexlab{b}}.
\newblock Towards expert-level medical question answering with large language models.
\newblock \emph{arXiv preprint arXiv:2305.09617}.

\bibitem[{Su et~al.(2024)Su, Mo, Tiwari, Wang, Nie, and Simonsen}]{su2024mixture}
Zhan Su, Fengran Mo, Prayag Tiwari, Benyou Wang, Jian-Yun Nie, and Jakob~Grue Simonsen. 2024.
\newblock Mixture of experts using tensor products.
\newblock \emph{arXiv preprint arXiv:2405.16671}.

\bibitem[{Touvron et~al.(2023{\natexlab{a}})Touvron, Lavril, Izacard, Martinet, Lachaux, Lacroix, Rozi{\`e}re, Goyal, Hambro, Azhar et~al.}]{touvron2023llama}
Hugo Touvron, Thibaut Lavril, Gautier Izacard, Xavier Martinet, Marie-Anne Lachaux, Timoth{\'e}e Lacroix, Baptiste Rozi{\`e}re, Naman Goyal, Eric Hambro, Faisal Azhar, et~al. 2023{\natexlab{a}}.
\newblock Llama: Open and efficient foundation language models.
\newblock \emph{arXiv preprint arXiv:2302.13971}.

\bibitem[{Touvron et~al.(2023{\natexlab{b}})Touvron, Martin, Stone, Albert, Almahairi, Babaei, Bashlykov, Batra, Bhargava, Bhosale et~al.}]{touvron2023llama2}
Hugo Touvron, Louis Martin, Kevin Stone, Peter Albert, Amjad Almahairi, Yasmine Babaei, Nikolay Bashlykov, Soumya Batra, Prajjwal Bhargava, Shruti Bhosale, et~al. 2023{\natexlab{b}}.
\newblock Llama 2: Open foundation and fine-tuned chat models.
\newblock \emph{arXiv preprint arXiv:2307.09288}.

\bibitem[{Van~Veen et~al.(2024)Van~Veen, Van~Uden, Blankemeier, Delbrouck, Aali, Bluethgen, Pareek, Polacin, Reis, Seehofnerov{\'a} et~al.}]{van2024adapted}
Dave Van~Veen, Cara Van~Uden, Louis Blankemeier, Jean-Benoit Delbrouck, Asad Aali, Christian Bluethgen, Anuj Pareek, Malgorzata Polacin, Eduardo~Pontes Reis, Anna Seehofnerov{\'a}, et~al. 2024.
\newblock Adapted large language models can outperform medical experts in clinical text summarization.
\newblock \emph{Nature Medicine}, pages 1--9.

\bibitem[{Wang et~al.(2023{\natexlab{a}})Wang, Liu, Xi, Qiang, Zhao, Qin, and Liu}]{wang2023huatuo}
Haochun Wang, Chi Liu, Nuwa Xi, Zewen Qiang, Sendong Zhao, Bing Qin, and Ting Liu. 2023{\natexlab{a}}.
\newblock Huatuo: Tuning llama model with chinese medical knowledge.
\newblock \emph{arXiv preprint arXiv:2304.06975}.

\bibitem[{Wang et~al.(2023{\natexlab{b}})Wang, Liu, Zhao, Qin, and Liu}]{ChatGLM-Med}
Haochun Wang, Chi Liu, Sendong Zhao, Bing Qin, and Ting Liu. 2023{\natexlab{b}}.
\newblock Chatglm-med: Chatglm model fine-tuning based on chinese medical knowledge.
\newblock \url{https://github.com/SCIR-HI/Med-ChatGLM}.

\bibitem[{Wang et~al.(2023{\natexlab{c}})Wang, Chen, Ge, Xia, Bao, Zheng, Zhang, Gui, and Huang}]{wang-etal-2023-orthogonal}
Xiao Wang, Tianze Chen, Qiming Ge, Han Xia, Rong Bao, Rui Zheng, Qi~Zhang, Tao Gui, and Xuanjing Huang. 2023{\natexlab{c}}.
\newblock \href {https://doi.org/10.18653/v1/2023.findings-emnlp.715} {Orthogonal subspace learning for language model continual learning}.
\newblock In \emph{Findings of the Association for Computational Linguistics: EMNLP 2023}, pages 10658--10671, Singapore. Association for Computational Linguistics.

\bibitem[{Wang et~al.(2023{\natexlab{d}})Wang, Chen, Song, Zhang, Chen, Xiao, Jiang, Li, Wan, Wang et~al.}]{wang2023cmb}
Xidong Wang, Guiming~Hardy Chen, Dingjie Song, Zhiyi Zhang, Zhihong Chen, Qingying Xiao, Feng Jiang, Jianquan Li, Xiang Wan, Benyou Wang, et~al. 2023{\natexlab{d}}.
\newblock Cmb: A comprehensive medical benchmark in chinese.
\newblock \emph{arXiv preprint arXiv:2308.08833}.

\bibitem[{Wei et~al.(2021)Wei, Bosma, Zhao, Guu, Yu, Lester, Du, Dai, and Le}]{wei2021finetuned}
Jason Wei, Maarten Bosma, Vincent~Y Zhao, Kelvin Guu, Adams~Wei Yu, Brian Lester, Nan Du, Andrew~M Dai, and Quoc~V Le. 2021.
\newblock Finetuned language models are zero-shot learners.
\newblock \emph{arXiv preprint arXiv:2109.01652}.

\bibitem[{Wu et~al.(2023)Wu, Yao, Chen, Pan, Wang, Liu, and Yu}]{wu2023language}
Xuansheng Wu, Wenlin Yao, Jianshu Chen, Xiaoman Pan, Xiaoyang Wang, Ninghao Liu, and Dong Yu. 2023.
\newblock From language modeling to instruction following: Understanding the behavior shift in llms after instruction tuning.
\newblock \emph{arXiv preprint arXiv:2310.00492}.

\bibitem[{Xiong et~al.(2023)Xiong, Wang, Zhu, Zhao, Liu, Huang, Wang, and Shen}]{xiong2023doctorglm}
Honglin Xiong, Sheng Wang, Yitao Zhu, Zihao Zhao, Yuxiao Liu, Linlin Huang, Qian Wang, and Dinggang Shen. 2023.
\newblock Doctorglm: Fine-tuning your chinese doctor is not a herculean task.
\newblock \emph{arXiv preprint arXiv:2304.01097}.

\bibitem[{Xu(2023)}]{MedicalGPT}
Ming Xu. 2023.
\newblock Medicalgpt: Training medical gpt model.
\newblock \url{https://github.com/shibing624/MedicalGPT}.

\bibitem[{Yang et~al.(2023)Yang, Xiao, Wang, Zhang, Bian, Yin, Lv, Pan, Wang, Yan et~al.}]{yang2023baichuan}
Aiyuan Yang, Bin Xiao, Bingning Wang, Borong Zhang, Ce~Bian, Chao Yin, Chenxu Lv, Da~Pan, Dian Wang, Dong Yan, et~al. 2023.
\newblock Baichuan 2: Open large-scale language models.
\newblock \emph{arXiv preprint arXiv:2309.10305}.

\bibitem[{Zeng et~al.(2022)Zeng, Liu, Du, Wang, Lai, Ding, Yang, Xu, Zheng, Xia et~al.}]{zeng2022glm}
Aohan Zeng, Xiao Liu, Zhengxiao Du, Zihan Wang, Hanyu Lai, Ming Ding, Zhuoyi Yang, Yifan Xu, Wendi Zheng, Xiao Xia, et~al. 2022.
\newblock Glm-130b: An open bilingual pre-trained model.
\newblock \emph{arXiv preprint arXiv:2210.02414}.

\bibitem[{Zhang et~al.(2022{\natexlab{a}})Zhang, Chen, Bi, Liang, Li, Shang, Yin, Tan, Xu, Huang, Si, Ni, Xie, Sui, Chang, Zong, Yuan, Li, Yan, Zan, Zhang, Tang, and Chen}]{zhang-etal-2022-cblue}
Ningyu Zhang, Mosha Chen, Zhen Bi, Xiaozhuan Liang, Lei Li, Xin Shang, Kangping Yin, Chuanqi Tan, Jian Xu, Fei Huang, Luo Si, Yuan Ni, Guotong Xie, Zhifang Sui, Baobao Chang, Hui Zong, Zheng Yuan, Linfeng Li, Jun Yan, Hongying Zan, Kunli Zhang, Buzhou Tang, and Qingcai Chen. 2022{\natexlab{a}}.
\newblock \href {https://doi.org/10.18653/v1/2022.acl-long.544} {{CBLUE}: A {C}hinese biomedical language understanding evaluation benchmark}.
\newblock In \emph{Proceedings of the 60th Annual Meeting of the Association for Computational Linguistics (Volume 1: Long Papers)}, pages 7888--7915, Dublin, Ireland. Association for Computational Linguistics.

\bibitem[{Zhang et~al.(2022{\natexlab{b}})Zhang, Chen, Bi, Liang, Li, Shang, Yin, Tan, Xu, Huang et~al.}]{zhang2022cblue}
Ningyu Zhang, Mosha Chen, Zhen Bi, Xiaozhuan Liang, Lei Li, Xin Shang, Kangping Yin, Chuanqi Tan, Jian Xu, Fei Huang, et~al. 2022{\natexlab{b}}.
\newblock Cblue: A chinese biomedical language understanding evaluation benchmark.
\newblock In \emph{Proceedings of the 60th Annual Meeting of the Association for Computational Linguistics (Volume 1: Long Papers)}, pages 7888--7915.

\bibitem[{Zhang et~al.(2018)Zhang, Wu, He, Liu, and Su}]{zhang2018medical}
Xiao Zhang, Ji~Wu, Zhiyang He, Xien Liu, and Ying Su. 2018.
\newblock Medical exam question answering with large-scale reading comprehension.
\newblock In \emph{Proceedings of the AAAI conference on artificial intelligence}, volume~32.

\bibitem[{Zhao et~al.(2022)Zhao, Li, Wu, Hu, Chen, Wang, Ding, and Zhang}]{zhao2022medical}
Yu~Zhao, Yunxin Li, Yuxiang Wu, Baotian Hu, Qingcai Chen, Xiaolong Wang, Yuxin Ding, and Min Zhang. 2022.
\newblock Medical dialogue response generation with pivotal information recalling.
\newblock In \emph{Proceedings of the 28th ACM SIGKDD Conference on Knowledge Discovery and Data Mining}, pages 4763--4771.

\bibitem[{Zhu et~al.(2023)Zhu, Wang, Zheng, Chen, and Tang}]{zhu2023promptcblue}
Wei Zhu, Xiaoling Wang, Huanran Zheng, Mosha Chen, and Buzhou Tang. 2023.
\newblock Promptcblue: A chinese prompt tuning benchmark for the medical domain.
\newblock \emph{arXiv preprint arXiv:2310.14151}.

\end{thebibliography}

\appendix

\clearpage
\section{Related Works}
\paragraph{Bilingual medical large language models}
Large language models such as GPT-4~\citep{elmohamed}, PaLM~\citep{chowdhery2023palm} and LLaMA~\citep{touvron2023llama,touvron2023llama2} have achieved superior zero-shot performance across tasks and serve as interactive chatbots to interact with humans. 
However, trained on little medical-oriented data and Chinese data, these LLMs are not useful for medical conversations and consultations, especially for Chinese medical scenarios.
Therefore, a lot of work has been done to fine-tune base models on medical data to obtain large medical models.
Med-PaLM~\citep{singhal2022large} is grounded on Flan-PaLM~\citep{chung2022scaling,chowdhery2023palm} to encode clinical knowledge.
Med-PaLM2~\citep{singhal2023towards} as a successor, reduce the gap with doctors by fine-tuning in the medical data and using modern prompting strategies.
Apart from grounding on super-large language models such as PaLM or GPT, many works attempt to build medical-LLM on deployable sizes of LLMs, including 7B and 13B. 
DoctorGLM~\citep{xiong2023doctorglm}, ChatGLM-Med~\citep{ChatGLM-Med}, and Bianque-2~\citep{chen2023bianque} are all built on ChatGLM to support acceptable bilingual medical consultation.
Other work like MedicalGPT~\citep{MedicalGPT}, Huatuo~\citep{wang2023huatuo} and HuatuoGPT-\uppercase\expandafter{\romannumeral2}~\citep{chen2023huatuogpt} are built on LLaMA-series and enlarge the vocabulary to support Chinese conversations.


\paragraph{Parameter efficient fine-tuning}
Full fine-tuning effectively adapts base large language models to downstream tasks but also consumes significant computational resources with the increasing size of models and the number of tasks. To address this, Parameter-Efficient Fine-Tuning (PEFT) methods have been introduced. These methods freeze the base language models, modifying only a negligible number of parameters during the training phase, yet achieving similar or even superior performance with limited fine-tuning data.
Among these methods, Adapter-Tuning~\citep{rebuffi2017learning,houlsby2019parameter,lin-etal-2020-exploring,pfeiffer-etal-2021-adapterfusion} was the pioneering architecture that connected two additional projection layers to the pretrained language model. 
In addition to incorporating additional modules, Prefix-tuning~\citep{li-liang-2021-prefix} introduces learnable prefix tokens and prepends them before the input prompts. 
Differently, Prompt-Tuning~\citep{lester-etal-2021-power} utilizes learnable prompt tokens for each task, as a multi-task PEFT approach in NLU scenarios. 
Following these, P-Tuning~\citep{LIU2023} and P-Tuning-v2~\citep{liu-etal-2022-p} move away from explicit prompts and employ a prompt-generator to convert pseudo prompts into task prompts, allowing decoder-only models to also perform NLU tasks. 
However, these methods introduce additional priors and significant inference latency.
In contrast, Low-Rank Adaptation (LoRA)\citep{hu2022lora} and its variant, Weight-Decomposed Low-Rank Adaptation (DoRA)\citep{liu2024dora}, take a different approach. 
LoRA updates original parameters with two low-rank matrices without assuming any specific task or architecture, eliminating inference latency by merging back these two matrices to the original weight. 
DoRA extends this by incorporating weight decomposition, achieving performance comparable to full fine-tuning. 
Nonetheless, transferring these methods to multi-task learning scenarios without manual adjustments remains a challenge.



\begin{table*}[t]
\centering
\resizebox{\textwidth}{!}{%
\begin{tabular}{ccccc}
\toprule
\textbf{Type} & \textbf{Task} & \textbf{Description}                                                   & \textbf{Size}                & \textbf{Metrics} \\
\midrule
              & MedQA         & Chinese Mainland Medical License Exams (USMLE)                & 3,425         & Accuracy         \\
              & MMedBnech & Chinese subset of the Multilingual Medical Benchmark
 & 3,425 & Accuracy \\ 
              & CMB           & Comprehensive Multi-level Assessment for Medical Knowledge & 11,200        & Accuracy         \\
              & CMExam        & Chinese National Medical Licensing Examination             & 6,811         & Accuracy         \\
              & CMMLU$^\dagger$        & Chinese Massive Multitask Language Understanding           & 1,354         & Accuracy         \\
              & CEval$^\dagger$         & A Multi-Level Multi-Discipline Chinese Evaluation          & 41            & Accuracy         \\
              & PLE Pharmacy & Pharmacist Licensure Examination Pharmacy track & 480 & Accuracy \\
\multirow{-8}{*}{\begin{tabular}[c]{@{}c@{}}Knowledge-Intensive \\ Tasks\end{tabular}}              & PLE TCM & Pharmacist Licensure Examination Traditional Chinese Medicine track & 480 & Accuracy \\
\midrule
    & Report Generation & Analysis of Abnormal Indicators in Physical Examination Reports & 50 & GPT-4 Eval \\
     & Image Analysis & Analysis of the Medical Image Reports & 50 & GPT-4 Eval \\ 
     & Discharge Instruction &  Providing patients with clear and comprehensive guidance  & 50 & GPT-4 Eval \\
\multirow{-4}{*}{\begin{tabular}[c]{@{}c@{}}CCTE\end{tabular}}      & Examination Education & Guide students in understanding the thought process behind examination questions & 50 & GPT-4 Eval \\
\midrule
              & CMeEE         & Chinese Medical Named Entity Recognition                               & 500                         & Micro-F1         \\
              & CMeIE         & Chinese Medical Text Entity Relationship Extraction                    & 600 & Micro-F1         \\
              & CHIP-CDN      & Clinical Terminology Normalization                                     & 600                        & Micro-F1         \\
              & CHIP-CDEE     & Clinical Discovery Event Extraction                                    & 600                          & Micro-F1         \\
              & IMCS-V2-NER   & Intelligent Medical Conversation System Named Entity Recognition       & 600                          & Micro-F1         \\
 &
  CHIP-MDCFNPC &
  Medical Dialog Clinical Findings Positive and Negative Classification &
  600 &
  Micro-F1 \\
              & IMCS-V2-SR    & Intelligent Medical Conversation System Symptom Recognition            & 600                          & Micro-F1         \\
              & IMCS-V2-DAC   & Intelligent Medical Conversation System Dialogue Action Classification & 800                          & Macro-F1         \\
              & IMCS-V2-MRG   & Intelligent Medical Conversation System Medical Report Generation      & 600                          & RougeL           \\
              & CHIP-CTC      & Clinical Trial Criterion                                               & 1100                        & Micro-F1         \\
              & CHIP-STS      & Semantic Textual Similarity                                            & 600                        & Micro-F1         \\
              & KUAKE-IR      & Information Retrieval                                                  & 600                        & Micro-F1         \\
              & KUAKE-QIC     & Query Intent Criterion                                                 & 660                        & Micro-F1         \\
              & KUAKE-QQR     & Query Query Relevance                                                  & 600                        & Micro-F1         \\
              & KUAKE-QTR     & Query Title Relevance                                                  & 600                        & Micro-F1         \\
\multirow{-16}{*}{\begin{tabular}[c]{@{}c@{}}CBLUE\end{tabular}} &
  MedDG &
  Medical Dialog Generation &
  600 &
  RougeL \\
\bottomrule
\end{tabular}%
}
\caption{Statistics of the evaluated medical tasks. "$\dagger$" indicates that we only choose the questions related to the medical domain for the task.}
\label{tab: statistics}
\end{table*}

\section{Experiments Details}
\label{appendix: experiments}

\subsection{Implementation Detail}
\label{appendix: implementations}
For all sizes of \ming, we set the batch size to 128 and fine-tuned the models with 1 epoch for the miscellaneous knowledge aggregation step. The learning rate is 2e-4, with a \texttt{warmup\_ratio}=0.03 and the cosine learning schedule. The maximum length of the training sample is configured to 3072. We fine-tune the model with MKA stage for 1 epoch and DA stagê for 3 epochs. The orthogonal weight factor $\lambda=1$. For the configuration of the PEFT, the LoRA rank $r$ and $\alpha$ are fixed at 16 and 32, respectively. The number of the shared experts is set to 2, and the total number of mixture experts is set to 8, with 2 experts activated for each token during the training. 
We only adopt MoLoRA for the linear layers in the feed-forward network~(FFN) blocks and adopt normal LoRA for the linear layers in the attention blocks. All the experiments are conducted on 8$\times$A100 80G GPUs.

\subsection{Baseline Models}
We selected various models as baselines for comparing their performance on medical tasks. For the general open-sourced models, we choose Baichuan2-7B/13B-Chat~\citep{yang2023baichuan}, Qwen1.5-7B/14B-Chat~\citep{bai2023qwen}, ChatGLM2/3-6b~\citep{zeng2022glm}, LLaMA2-7B/13B-Chat~\citep{touvron2023llama}, and LLaMA3-8B-Instruct~\citep{metaIntroducingMeta}. We choose HuatuoGPT-II~(7B/13B)~\citep{chen2023huatuogpt} for the medical open-sourced models. Additionally, we also choose the ChatGPT~\citep{elmohamed} as the type of closed-source model with strong performance.

\subsection{Fine-tuning Corpus}
\label{appendix: training data}
The corpus used for the MKA fine-tuning stage contains nearly 400k samples in total. For the publicly available parts of data, it contains 80k multiple-choice of the question-answering sample with rational from the training set of the MMedBench~\citep{qiu2024towards} and CMExam~\citep{liu2024benchmarking}, and 68k samples from the PromptCBLUE~\citep{zhu2023promptcblue}, which transfer CBLUE~\citep{zhang-etal-2022-cblue} into a pure text format using specific prompts, 30k multi-turn medical conversations sampled from the \texttt{HuatuoGPT-sft-data-v1}~\citep{wang2023huatuo} and 50k single-turn question-answering data generated by GPT-4 from \texttt{HuatuoGPT2\_sft\_instruct\_GPT4\_50K}~\citep{chen2023huatuogpt}. The private portion of the data contains about 200k training samples, with each of the four clinical tasks having 50k training samples each.

\subsection{Testing Benchmarks}
\label{appendix: testing data}
A detailed summary of the descriptions, quantities, and evaluation methods of all test data can be found in Table~\ref{tab: statistics}.
\paragraph{Medical Knowledge Exams} Medical licensing exams measure large language models' medical knowledge and reasoning capabilities, serving as a common testing method. For this type of Benchmarks, we follow the experiments setting of~\citet{chen2023huatuogpt}. The examination benchmarks include the Chinese Mainland test set of MedQA~\citep{jin2021disease}, the Chinese test set of MMedBench~\citep{qiu2024towards}, and two comprehensive Chinese medical exam datasets, validation set of the CMB~\citep{wang2023cmb} and the test set of the CMExam~\citep{liu2024benchmarking}. We also collect the medical parts of the general benchmarks, including the validation set of the CMMLU~\citep{li2023cmmlu} and CEval~\citep{huang2023ceval}. We also test the performance of the models on the flash exam questions from the
2023 Chinese National Pharmacist Licensure Examination, collected by~\citet{chen2023huatuogpt}.

\paragraph{Medical Alignment Tasks} The characteristic of medical alignment Tasks lies in their unique input and output format requirements, primarily evaluating the models' ability to follow the instructions in the clinical scenario. We mainly choose two types of alignment tasks to evaluate the models. The first is CBLUE~\citep{zhang-etal-2022-cblue}, a Chinese multi-task medical dataset encompassing 16 distinct medical NLP tasks. The second is the proposed Chinese Clinical Task Evaluation~(CCTE), which comprises four clinical tasks: Report Generation, Image Analysis, Discharge Instruction, and Examination Education. 

\section{Chinese Clinical Tasks Evaluation}
\label{appendix: ccte}
\subsection{Tasks Descriptions} 

\begin{figure*}[thp]
    \centering
    \includegraphics[width=1.0\linewidth]{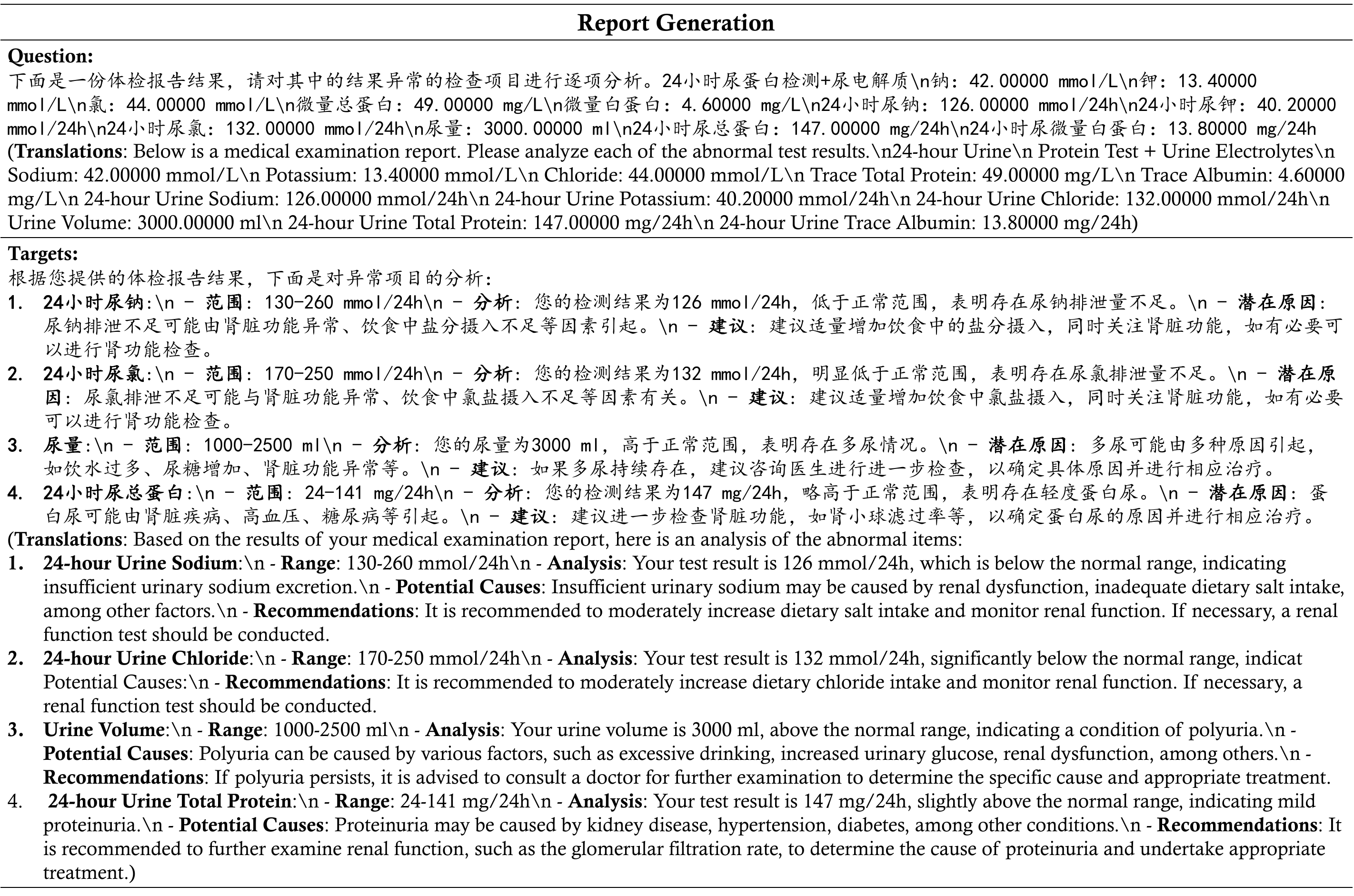}
    \caption{Case Example of Report Generation tasks.}
    \label{fig: report generation}
\end{figure*}

\paragraph{Report Generation} 
In the Report Generation task, there are over 600 different types of test items and more than 3,600 test reports containing various combinations of these test items. The input comprises patient-specific laboratory test report data. The model is tasked with analyzing and identifying which of the patient's test indicators deviate from normal ranges, providing insights into potential underlying causes and offering relevant recommendations for further action. Typically, the number of test items presented in the reports ranges from 2 to 16. To effectively accomplish this task, the model must demonstrate a robust capability for recognizing and interpreting extended contextual information, as well as possess an advanced understanding of medical knowledge. This is crucial for ensuring accurate analysis and the provision of actionable medical advice based on the test results. The data example is shown in Figure~\ref{fig: report generation}.

\begin{figure*}[thp]
    \centering
    \includegraphics[width=1.0\linewidth]{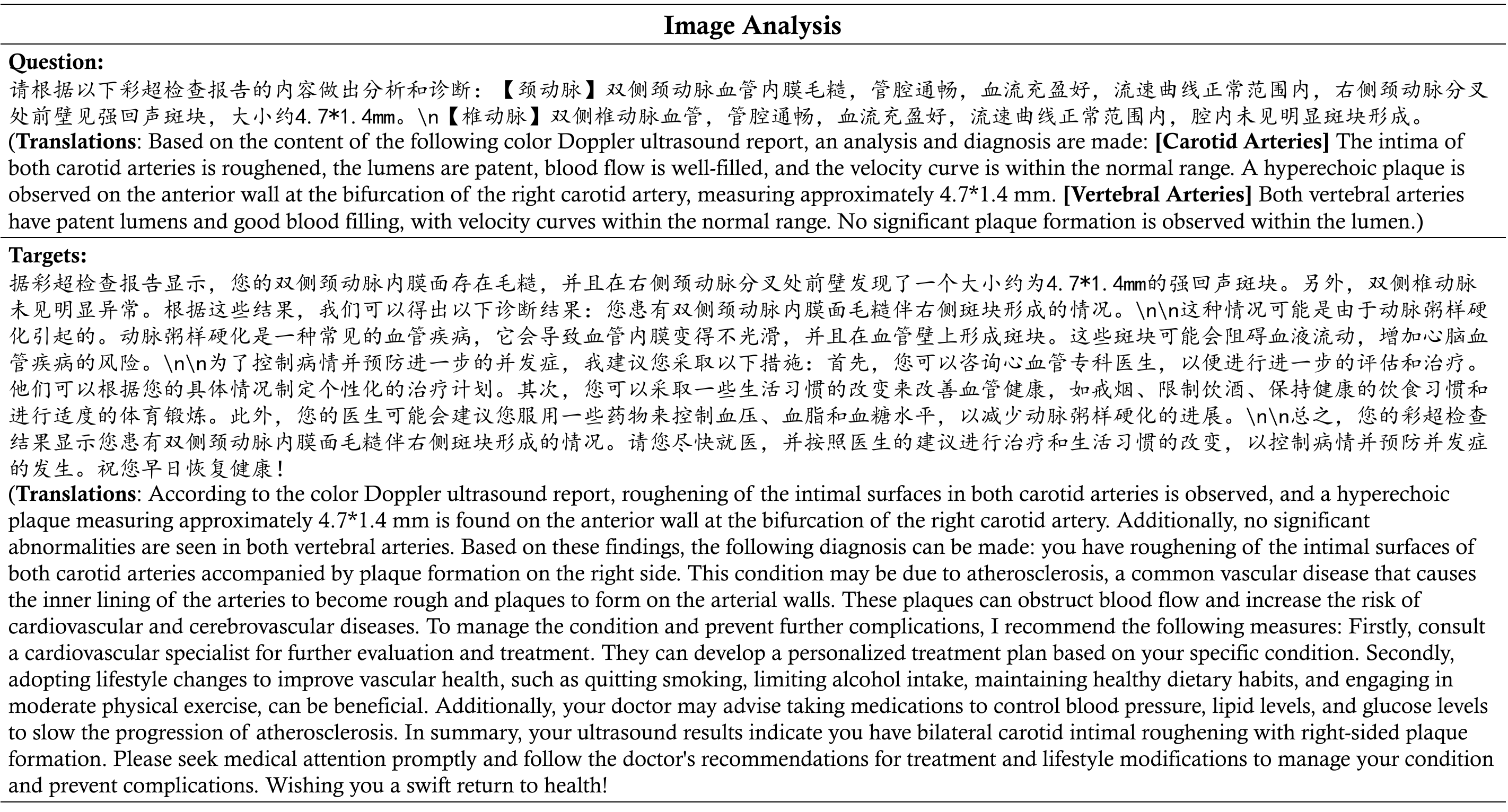}
    \caption{Case Example of Image Analysis tasks.}
    \label{fig: image analysis}
\end{figure*}

\paragraph{Image Analysis} 
In the Image Analysis task, the primary focus is on generating, analyzing, and diagnosing based on descriptive reports derived from medical imaging. This task involves interpreting detailed reports associated with 14 types of medical image reports, including Rapid Pathology, Endoscopy, Magnetic Resonance Imaging, Electrocardiogram, Computed Tomography, Color Ultrasound, Digital Subtraction Angiography, Computed Radiography, Routine Pathology, Nuclear Medicine, Gastrointestinal Pathology, Endoscopy Examination, Radio Frequency, and Immunohistochemistry Pathology. The model must effectively parse and understand these textual descriptions to identify any noted abnormalities, correlate them with potential medical conditions, and provide diagnostic insights. The data example is shown in Figure~\ref{fig: image analysis}.

\begin{figure*}[thp]
    \centering
    \includegraphics[width=1.0\linewidth]{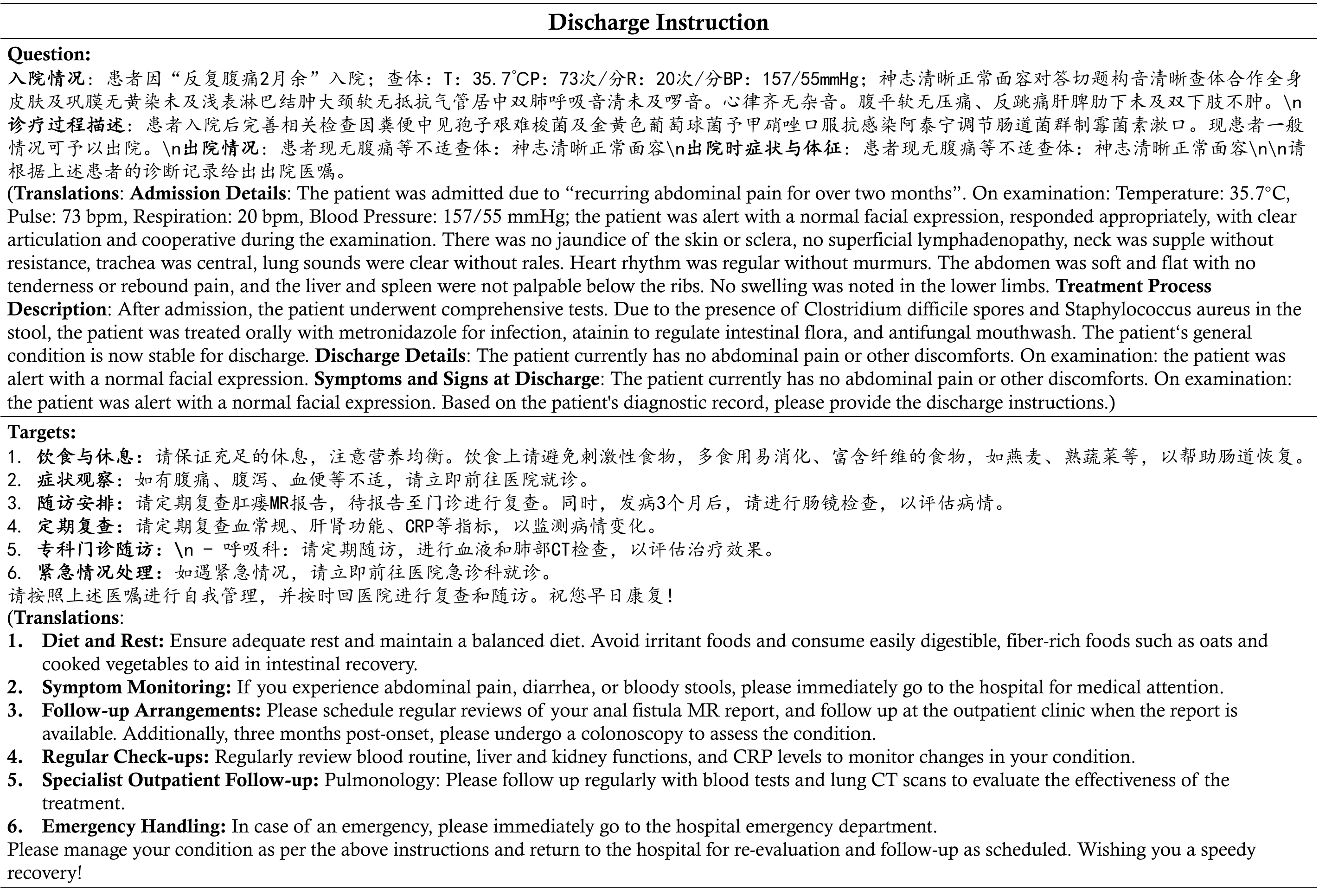}
    \caption{Case Example of Discharge Instruction tasks.}
    \label{fig: discharge instruction}
\end{figure*}

\paragraph{Discharge Instruction}
In the Discharge Instruction task, the primary objective is to generate comprehensive medical advice and instructions for patients at the time of their discharge based on their admission details, treatment processes, and discharge conditions. This task involves synthesizing information from a patient’s entire hospital stay, encompassing initial symptoms, diagnostic findings, treatments administered, and the patient’s response to those treatments. The model must adeptly process and integrate this wide array of medical data to formulate clear and precise discharge instructions. These instructions typically include guidelines on medication management, wound care, lifestyle adjustments, follow-up appointments, and signs of potential complications that should prompt immediate medical attention. The data example is shown in Figure~\ref{fig: discharge instruction}.

\begin{figure*}[thp]
    \centering
    \includegraphics[width=1.0\linewidth]{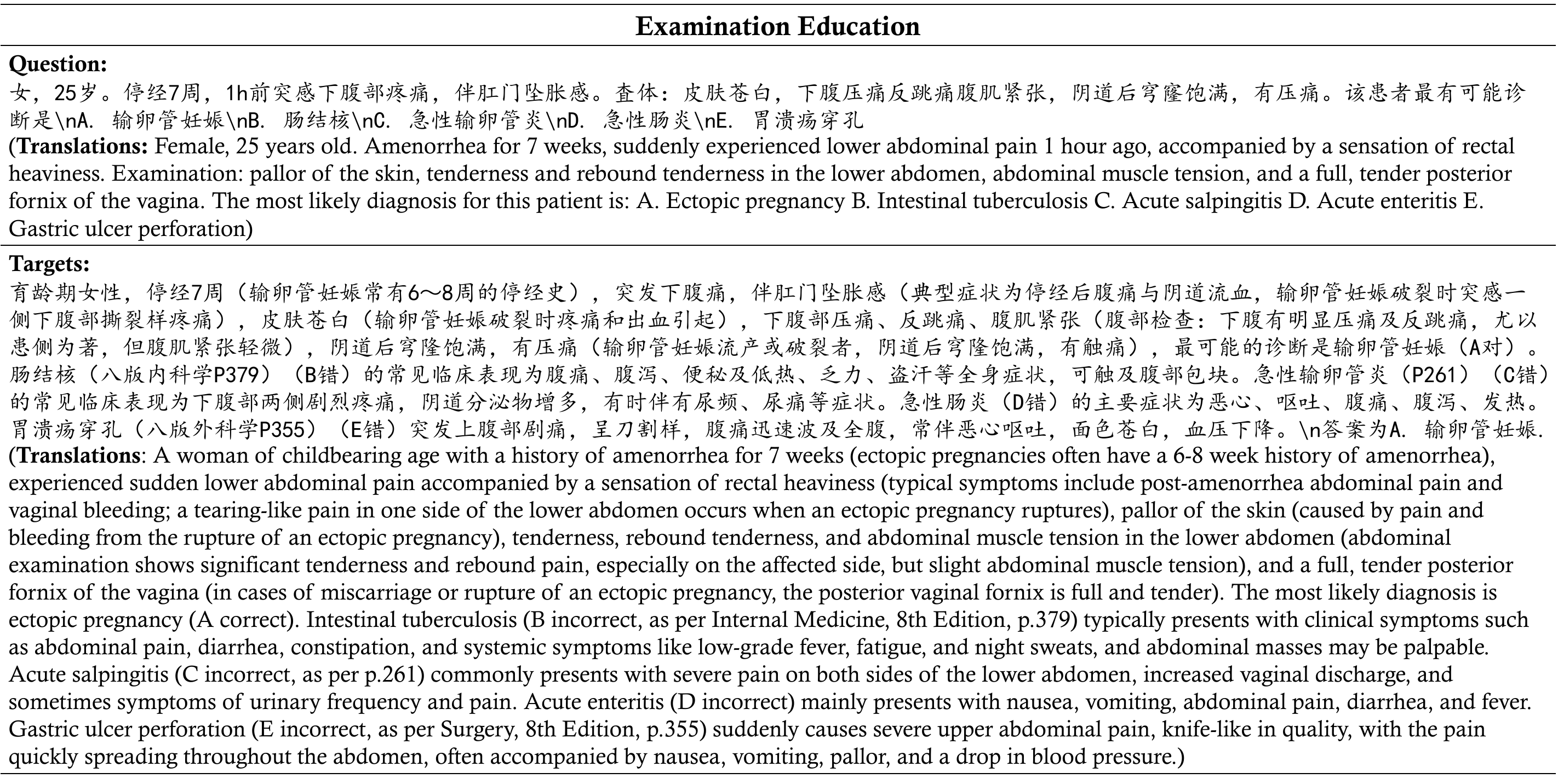}
    \caption{Case Example of Examination Education tasks.}
    \label{fig: examination education}
\end{figure*}

\paragraph{Examination Education}
The primary focus of the Examination Education task is on enhancing models' ability to explain their decision-making processes, particularly in medical contexts. This task involves evaluating the model's capability to provide detailed explanations and analyses of its responses. This not only aids in increasing the interpretability of large language models in healthcare applications but also serves as a valuable tool for medical students preparing for exams. For the data of the Examination Education, we sample the 50 samples from the test set of the MMedBench~\citep{qiu2024towards} for each question in it contains the ground truth rational. The data example is shown in Figure~\ref{fig: examination education}.

\begin{figure*}[t]
    \centering
    \includegraphics[width=1.0\linewidth]{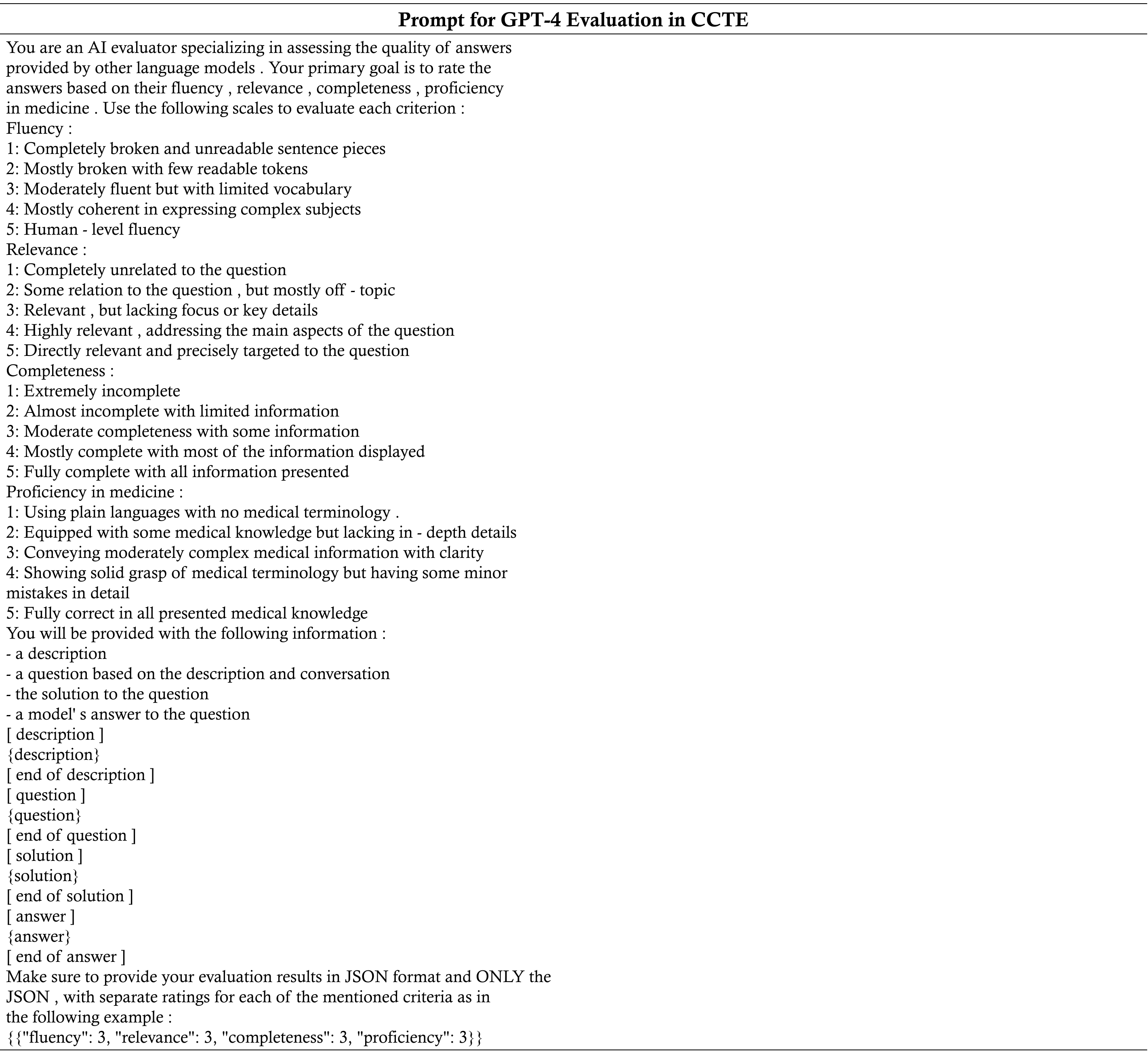}
    \caption{The prompt of the GPT-4 evaluation for CCTE tasks.}
    \label{fig: gpt4 eval prompt}
\end{figure*}

\subsection{Evaluation Method}
For the evaluation method of the CCTE dataset, we adopt the GPT4 evaluation to solve the problem that clinical tasks are difficult to evaluate. Following the evaluation pipeline of~\citet{wang2023cmb}, we score each response across four aspects — Fluency, Relevance, Completeness, and Medical Proficiency — using a grading scale from 1 to 5. We input the original data from clinical data for this task, along with the model's responses, into GPT-4 as evaluation criteria to prevent biases in scoring due to medical knowledge hallucinations in GPT-4. Simultaneously, GPT-4 leverages its capabilities to score language metrics such as fluency. The prompt of the GPT-4 is shown in Figure~\ref{fig: gpt4 eval prompt}.


\begin{table*}[]
\centering
\resizebox{\textwidth}{!}{%
\begin{tabular}{l|ccccc|ccccc|c}
\toprule
\multirow{2}{*}{\textbf{MODEL}} & \multicolumn{5}{c}{\textbf{Pharmacist Licensure Examination (Pharmacy)}} & \multicolumn{5}{c}{\textbf{Pharmacist Licensure Examination (TCM)}} & \multirow{2}{*}{\textbf{AVERAGE}} \\
 & \textbf{\begin{tabular}[c]{@{}c@{}}Optimal\\ Choice\end{tabular}} & \textbf{\begin{tabular}[c]{@{}c@{}}Matched\\ Selection\end{tabular}} & \textbf{\begin{tabular}[c]{@{}c@{}}Integrated\\ Analysis\end{tabular}} & \textbf{\begin{tabular}[c]{@{}c@{}}Multiple\\ Choice\end{tabular}} & \textbf{\begin{tabular}[c]{@{}c@{}}Total\\ Score\end{tabular}} & \textbf{\begin{tabular}[c]{@{}c@{}}Optimal\\ Choice\end{tabular}} & \textbf{\begin{tabular}[c]{@{}c@{}}Matched\\ Selection\end{tabular}} & \textbf{\begin{tabular}[c]{@{}c@{}}Integrated\\ Analysis\end{tabular}} & \textbf{\begin{tabular}[c]{@{}c@{}}Multiple\\ Choice\end{tabular}} & \textbf{\begin{tabular}[c]{@{}c@{}}Total\\ Score\end{tabular}} &  \\
 \midrule
Llama2-7B & 16.88 & 24.09 & 15.00 & 0.00 & 18.54 & 15.63 & 17.27 & 21.67 & 2.50 & 16.04 & 17.29 \\
Llama2-13B & 15.63 & 26.82 & 15.00 & 0.00 & 19.38 & 18.13 & 23.18 & 21.67 & 0.00 & 19.38 & 19.38 \\
Llama3-8B & 41.25 & 42.72 & 30.00 & 15.00 & 38.33 & 38.13 & 30.90 & 35.00 & 27.50 & 33.54 & 35.94 \\
ChatGLM2-6B & 37.00 & 36.80 & 25.00 & 31.70 & 35.30 & 33.10 & 37.30 & 35.00 & 37.30 & 33.70 & 34.50 \\
ChatGLM3-6B & 39.50 & 39.10 & 10.50 & 0.20 & 34.60 & 31.80 & 38.20 & 25.00 & 20.00 & 32.90 & 33.75 \\
Biachuan2-7B & 51.20 & 50.90 & 30.00 & 2.60 & 44.60 & 48.10 & 46.00 & 35.00 & 7.50 & 42.10 & 43.35 \\
Biachuan2-13B & 43.80 & 52.70 & 36.70 & 7.90 & 44.20 & 41.30 & 46.40 & 43.30 & 15.00 & 41.70 & 42.95 \\
Qwen1.5-7B & 29.38 & 37.73 & 30.00 & 17.50 & 32.29 & 30.63 & 34.09 & 18.33 & 12.50 & 29.17 & 30.73 \\
Qwen1.5-14B & 53.75 & 51.36 & 46.67 & 12.50 & 48.33 & 40.00 & 37.73 & 43.33 & 27.50 & 38.33 & 43.33 \\
ChatGPT & 45.60 & 44.10 & 36.70 & 13.20 & 41.20 & 34.40 & 32.30 & 30.00 & 15.00 & 31.20 & 36.20 \\
HuatuoGPT-II-7B & 41.90 & 61.00 & 35.00 & 15.70 & 47.70 & 52.50 & 51.40 & 41.70 & 15.00 & 47.50 & 47.60 \\
HuatuoGPT-II-13B & 47.50 & 64.10 & 45.00 & 23.70 & 52.90 & 48.80 & 61.80 & 45.00 & 17.50 & 51.60 & 52.25 \\
GPT-4 & 66.88 & 65.91 & 46.67 & 40.00 & 61.67 & 39.38 & 50.45 & 45.00 & 32.50 & 44.58 & 53.13 \\
\midrule
\ming-1.8B & 33.75 & 42.73 & 18.33 & 0.25 & 33.33 & 40.00 & 40.91 & 31.67 & 12.50 & 37.08 & 35.21 \\
\hspace{1em} $\mapsto$ \emph{w/o} DA & 34.38 & 44.55 & 30.00 & 5.00 & 36.04 & 40.00 & 42.73 & 45.00 & 12.50 & 39.58 & 37.81 \\
\ming-7B & 51.88 & 53.18 & 40.00 & 17.50 & 48.13 & 48.75 & 59.09 & 51.67 & 25.00 & 51.88 & 50.00 \\
\hspace{1em} $\mapsto$ \emph{w/o} DA & 50.63 & 60.91 & 38.33 & 15.00 & 50.83 & 52.50 & 62.27 & 38.33 & 30.00 & 53.33 & 52.08 \\
\ming-14B & 53.75 & 60.91 & 40.00 & 20.00 & 52.50 & 51.88 & 58.64 & 48.33 & 32.50 & 52.92 & 52.71 \\
\hspace{1em} $\mapsto$ \emph{w/o} DA & 59.38 & 67.27 & 36.67 & 25.00 & 57.29 & \textbf{56.88} & 57.73 & 51.67 & 27.50 & \textbf{54.17} & \textbf{55.73} \\
\bottomrule
\end{tabular}%
}
\caption{Results of the 2023 Chinese National Pharmacist Licensure Examination. It consists of
two separate Examinations including Pharmacy track and Traditional Chinese Medicine (TCM)
Pharmacy track. The results of the baseline models are obtained from~\citet{chen2023huatuogpt}}
\label{tab: ple detail results}
\end{table*}

\end{document}